\documentclass[11pt, a4paper, onecolumn, copyright, goog]{handshaketemplate}

\usepackage{multirow}
\usepackage[authoryear, sort&compress, round]{natbib}
\usepackage{tikz}
\usetikzlibrary{calc}
\usepackage{float}
\usepackage[svgnames, table]{xcolor}
\definecolor{crimson}{RGB}{220,20,60}
\usepackage[most]{tcolorbox}
\usepackage{tabularx}
\usepackage{booktabs}
\usepackage{wrapfig}
\usepackage{caption}
\usepackage{placeins}
\usepackage{enumitem}
\usepackage{parskip}
\usepackage{tcolorbox}
\tcbuselibrary{listingsutf8, breakable, skins}
\usepackage{array}
\usepackage{pifont}    % for ding symbols (e.g. \cmark, \xmark)
\usepackage{fvextra}   % breakable verbatim
\usepackage{wrapfig}
\usepackage{hyperref}
\usepackage{url}

\usepackage[nameinlink,compress]{cleveref}
\AtBeginDocument{\renewcommand{\ref}[1]{\errmessage{Use \noexpand\Cref instead of \noexpand\ref}}}

\usepackage{graphicx}

\newcommand{\cornerlogo}[2][1]{%
  \begin{tikzpicture}[remember picture,overlay]
    \node[anchor=north west, xshift=2.0cm, yshift=-1.8cm] at (current page.north west)
      {\includegraphics[scale=#1]{#2}};
    \draw[line width=0.3pt]
      ($(current page.north west)+(2.0cm,-2.7cm)$) -- ($(current page.north east)+(-2.0cm,-2.7cm)$);
  \end{tikzpicture}%
}

\usepackage{amsmath,amsfonts,bm}

\def\eqref#1{equation~\ref{#1}}
\def\1{\bm{1}}

\DeclareMathAlphabet{\mathsfit}{\encodingdefault}{\sfdefault}{m}{sl}
\SetMathAlphabet{\mathsfit}{bold}{\encodingdefault}{\sfdefault}{bx}{n}

\newcommand{\numexpertreviews}{161}
\newcommand{\numexpertreviewers}{26}
\newcommand{\academiashare}{62\%}
\newcommand{\industryshare}{35\%}
\newcommand{\sixyearshare}{69\%}
\newcommand{\industryexperienceshare}{88\%}

\newcommand{\professionalshare}{63\%}

\newcommand{\hleprofessionalshare}{22\%}

\newcommand{\projectnameweeklydailyshare}{33\%}
\newcommand{\hleweeklydailyshare}{14\%}

\newcommand{\commercialshare}{81\%}
\newcommand{\hlecommercialshare}{66\%}
\newcommand{\projectname}{VIALS}
\newcommand{\datasetsize}{161}

\newcommand{\bestmodel}{GPT-5.6 Sol and Gemini 3.7 Flash}
\newcommand{\bestscore}{26.5\% accuracy}
\title{VIALS: A Benchmark for Visual Interpretation of Artifacts in the Life Sciences}

\renewcommand{\today}{}

\makeatletter
\renewcommand\AB@authnote[1]{\textsuperscript{#1}}
\renewcommand\AB@affilnote[1]{\textsuperscript{#1}}
\makeatother

\makeatletter
\renewcommand\AB@authnote[1]{}
\renewcommand\AB@affilnote[1]{}
\makeatother

\author[]{Elaine Lau}
\author[]{Thanuka Udumulla}
\author[]{Lee Izhaki-Tavor}
\author[]{Francisco Guzmán}
\author[]{Nicholas Magazine}
\author[]{Jonas Mueller}
\affil[]{Handshake AI}

\begin{abstract}
In professional life sciences workflows, scientists routinely interpret visual artifacts (gel blots, microscopy images, plasmid maps, flow cytometry plots, molecular structures, ...) to inform research decisions.
We introduce VIALS, a visual question-answering benchmark with \datasetsize{} such interpretation tasks, spanning the types of artifacts examined throughout experimental workflows in the biotech industry (rather than polished figures from publications and textbooks).
While frontier vision-language models can now fluently describe natural images, we find that they are unable to accurately interpret these scientific images, reflecting limitations in domain knowledge and domain-specific visual reasoning capabilities.
In contrast, scientists with relevant domain expertise find these visual interpretation tasks straightforward. 
AI that cannot similarly interpret such images will have limited utility in professional life sciences workflows, where such artifacts are central to how scientists reason, communicate, and make decisions.
\end{abstract}

\begin{document}

\cornerlogo[0.27]{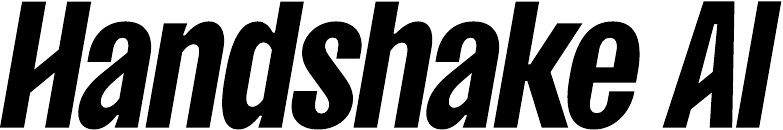}

\maketitle

\begingroup
\renewcommand{\thefootnote}{}
\makeatletter
\renewcommand{\@makefntext}[1]{%
  \noindent\hspace*{1.5em}%
  \begin{minipage}[t]{\dimexpr\linewidth-1.5em\relax}
  #1
  \end{minipage}%
}
\makeatother
\footnotetext{%
The benchmark dataset is available at \href{https://huggingface.co/datasets/Handshake-AI-Research/VIALS}{huggingface.co/datasets/Handshake-AI-Research/VIALS}\par
Code to run the benchmark is available at \href{https://github.com/Handshake-AI-Research/VIALS}{github.com/Handshake-AI-Research/VIALS}\par
Correspondence to \{elaine.lau, nicholas.magazine, jonas.mueller\}@joinhandshake.com}
\endgroup

\section{Introduction}
In the life sciences domain, information critical to research workflows is often encoded in visual artifacts such as gel blots, microscopy images, plasmid maps, phylogenetic trees, flow cytometry plots, docking views, and molecular structures. 
Current vision-language models (VLM) can generate reasonable descriptions of images, yet such descriptions do not guarantee an accurate expert interpretation of the critical information that enables scientific progress \citep{fu2024eccv-blink,yue2024mmmupro,groundedcot}. 
Interpreting these scientific visual artifacts requires models to localize the relevant evidence, extract the correct labels or values, distinguish signal from noise, compare visual elements, and reason about the information---all of which require proper application of the underlying scientific concepts.

A model can fail at any of these steps even when it recognizes the artifact and understands the underlying biology. It may identify the wrong band in a gel lane, misread a flow cytometry gate, or overlook a relevant structural feature in a complex protein-ligand interaction. These failures point to capability shortcomings that go beyond text recognition: they involve localization, comparison, and interpretation of spatial or structural relationships. A further concern is that the final answer may still sound fluent and scientifically plausible, making errors difficult to detect. Previous biomedical evaluations have identified visual perception and grounding as sources of failure \citep{microvqa,gpt4vmedical}.  To progress toward more economically valuable AI for the biotech industry, it is thus important to assess: \emph{How accurately can vision-language models interpret the scientific images routinely encountered in professional life sciences workflows?}

\begin{figure*}[h]
    \vspace{-1em}
    \centering
    \includegraphics[width=0.9\textwidth]{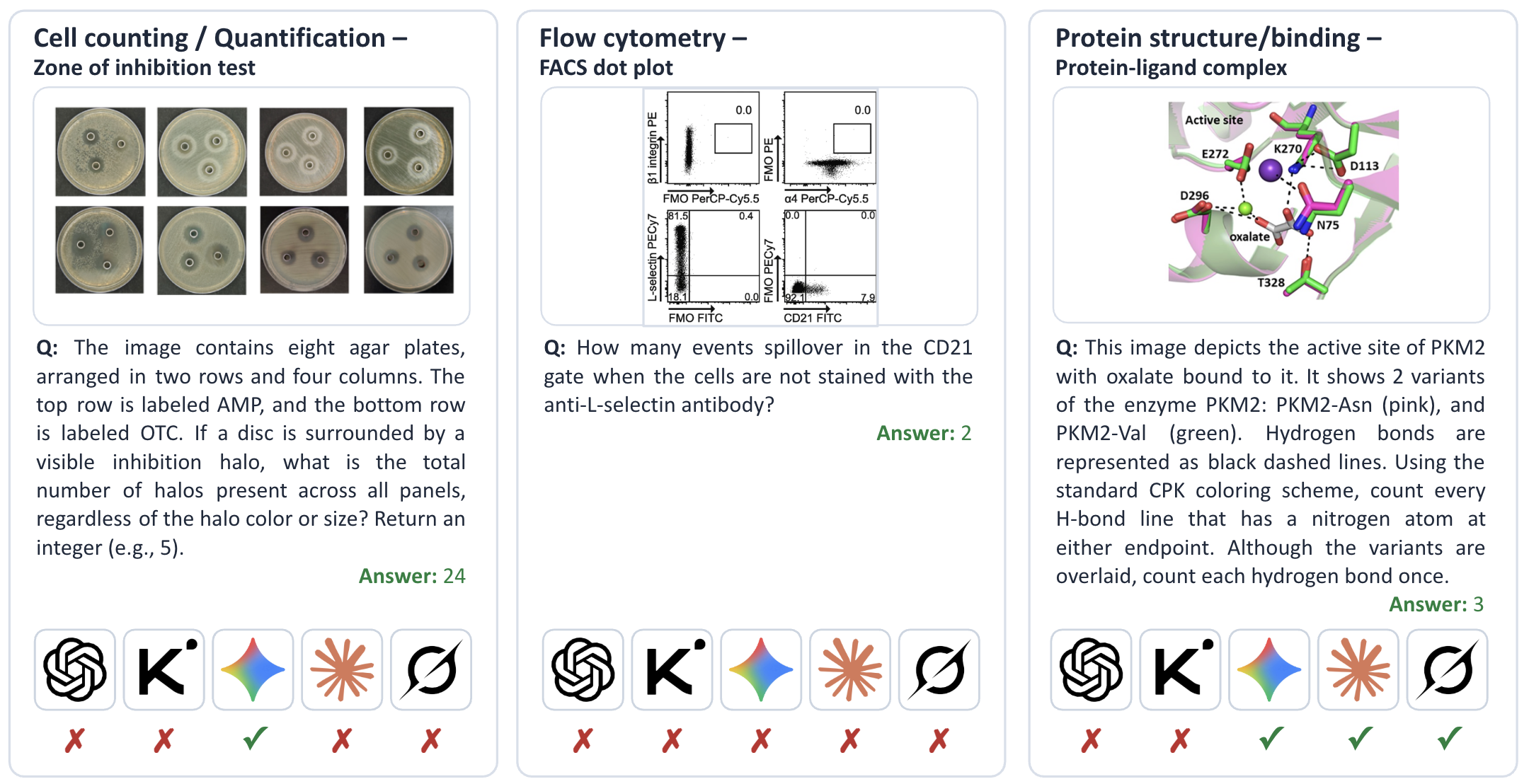}
    \captionsetup{font=small}
    \caption{Three tasks from the \projectname{} benchmark. (\emph{Left}) The \textbf{Zone of inhibition} artifacts depict agar plates containing growing microorganisms, and this task requires identifying the clear zones around the disks that reveal antibiotic activity---a common assessment in antibiotic discovery and clinical microbiology. (\emph{Middle}) The \textbf{FACS dot plot} artifacts visualize the results of a flow cytometry assay to characterize cell populations, and this task requires identifying cell detection events for a particular biomarker---a key step in immunology research and cell therapy development.  (\emph{Right}) The \textbf{Protein-ligand complex} artifact is a 3D visualization from structural biology software, showing how a molecule fits into a protein, and this task requires identifying the molecular contacts that enable binding---a key step in structure-based drug discovery.
    For each task, the correctness of answers from various models is shown along the bottom (GPT-5.6 Sol, Kimi K3, Gemini 3.1 Pro, Claude Opus 5, and Grok 4.6, from left to right). 
    }
    \label{fig:examples}
    \vspace{-1em}
\end{figure*}

Existing benchmarks poorly cover this area, focusing on general academic knowledge \citep{phan2025lastexam,yue2024mmmu,yue2024mmmupro,scivqa} or solely on specialized biomedical modalities such as microscopy, pathology, or clinical imaging \citep{pathvqa,microvqa,mmbu,gmaimmb}.
We introduce \projectname{}, a benchmark for \textbf{v}isual \textbf{i}nterpretation of \textbf{a}rtifacts in professional \textbf{l}ife \textbf{s}ciences. \projectname{} contains \datasetsize{} visual question-answering (VQA) tasks spanning many industry-relevant scientific domains and artifact types. Each task pairs a scientific image with a question requiring the extraction and interpretation of visual evidence central to  professional research workflows. The tasks are created and reviewed by PhD-level professional scientists with relevant domain and work expertise, ensuring they faithfully reflect real research workflows and have correct answers (\Cref{sec:benchmark_description}).
\Cref{fig:examples} shows representative benchmark examples.

For scientists trained in the relevant domains, many \projectname{} tasks are everyday parts of scientific practice, and performing them accurately is critical for research progress. A model that cannot reliably perform these tasks cannot practically be trusted in high-stakes life sciences workflows.
We evaluate all of today's best available multimodal models, observing that the top-performing models, \bestmodel{}, both achieve only \bestscore{}. Around 90\% of unsuccessful tasks involve errors in reading the artifact correctly, including miscounting, misreading measurements, or missing spatial and structural relationships (\Cref{sec:error-analysis}). Our results show that reliable interpretation of scientific artifacts remains a major bottleneck for current multimodal models, which struggle to properly apply scientific domain knowledge during visual processing and reasoning.

\section{Related Work} \label{sec:background} 

% \vspace*{-5mm}
\begin{figure}[!t]
% \vspace*{-5mm}
    \centering
    \begin{minipage}[t]{ \textwidth}
    % \vspace*{-5mm}
    \includegraphics[width=\textwidth]{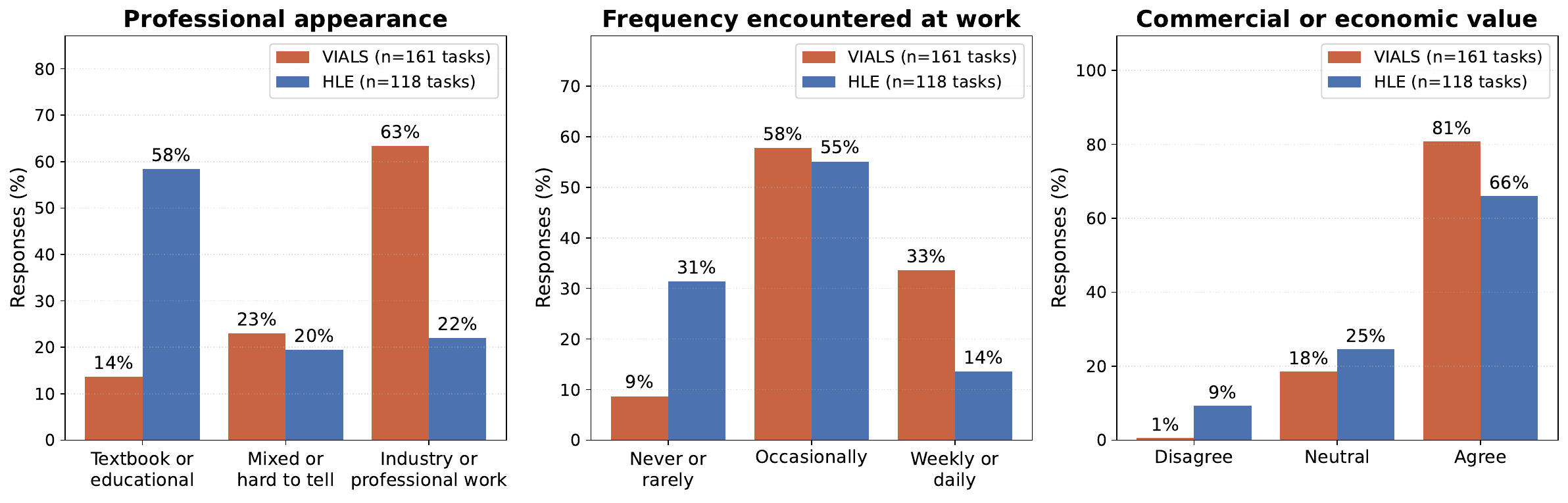}
        \vspace{-1.8em}
    \captionsetup{font=small}
    \captionof{figure}{
    Results from an expert survey (detailed in \Cref{sec:expert-assessment}) in which scientists with relevant domain expertise rated the professional relevance of individual benchmark tasks from \projectname{} and HLE's Chemistry and Biology/Medicine VQA domains. 
    Professional scientists assess whether each artifact appears to stem from professional work rather than educational material, how often they encounter this type of artifact in their work, and whether this interpretation task has commercial or economic value.
    }
    \label{fig:expert-assessment}
    \end{minipage}
    
    \begin{minipage}[t]{ \textwidth}
    \centering
    \begin{subfigure}[t]{0.48\linewidth}
        \centering
        \includegraphics[width=\linewidth]{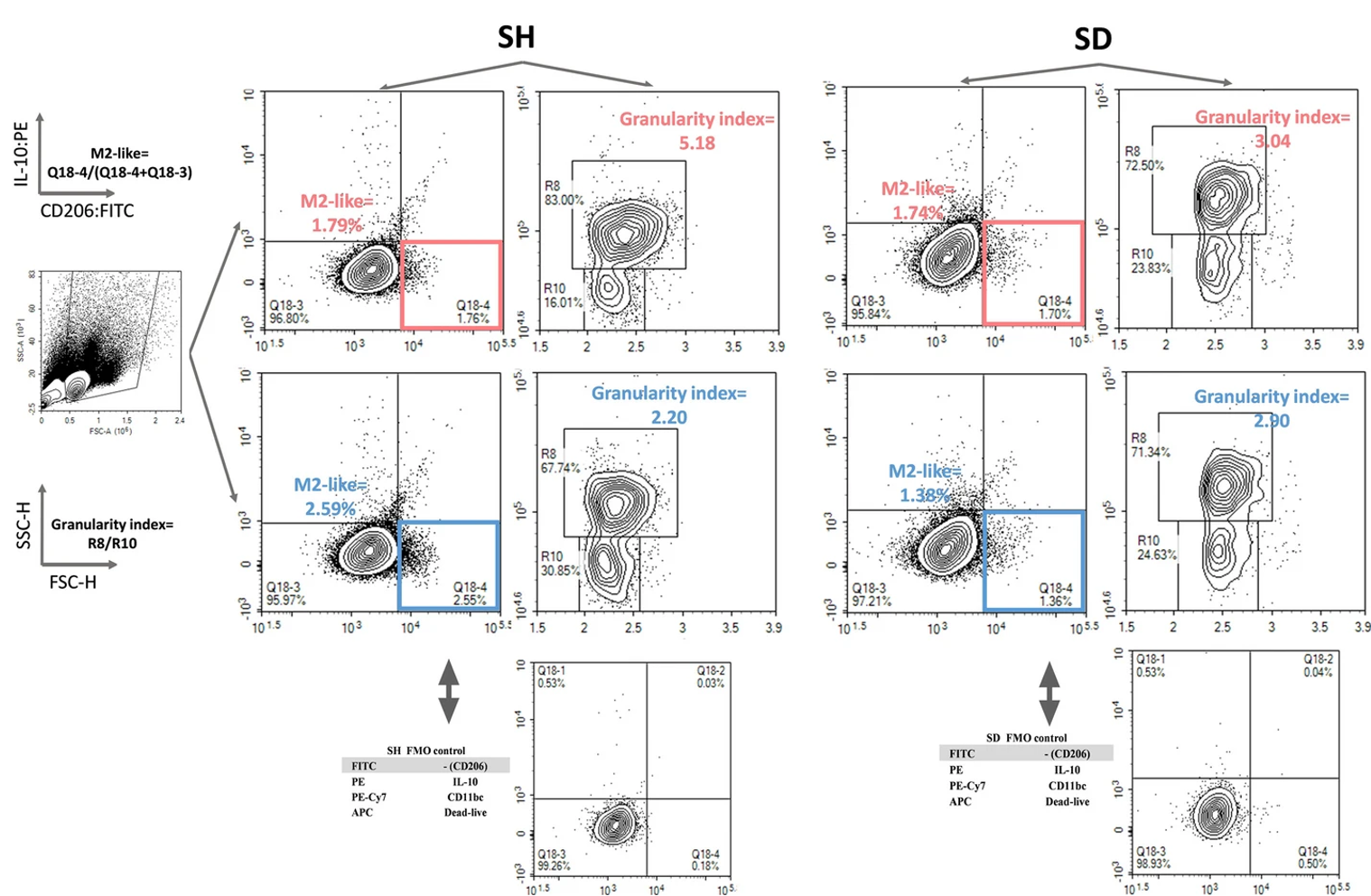}
        \captionsetup{font=small}
        \caption{
        \textbf{Flow cytometry task in \projectname{}.}
        \textit{Q.} What is the largest minus the smallest background-corrected high-granularity M2-like yield per 100,000 parent events? Round to the nearest whole cell.\quad
        \textit{A.} 989 cells per 100,000 parent events
        }
    \end{subfigure}
    \hfill
    \begin{subfigure}[t]{0.48\linewidth}
        \centering
        \includegraphics[width=\linewidth]{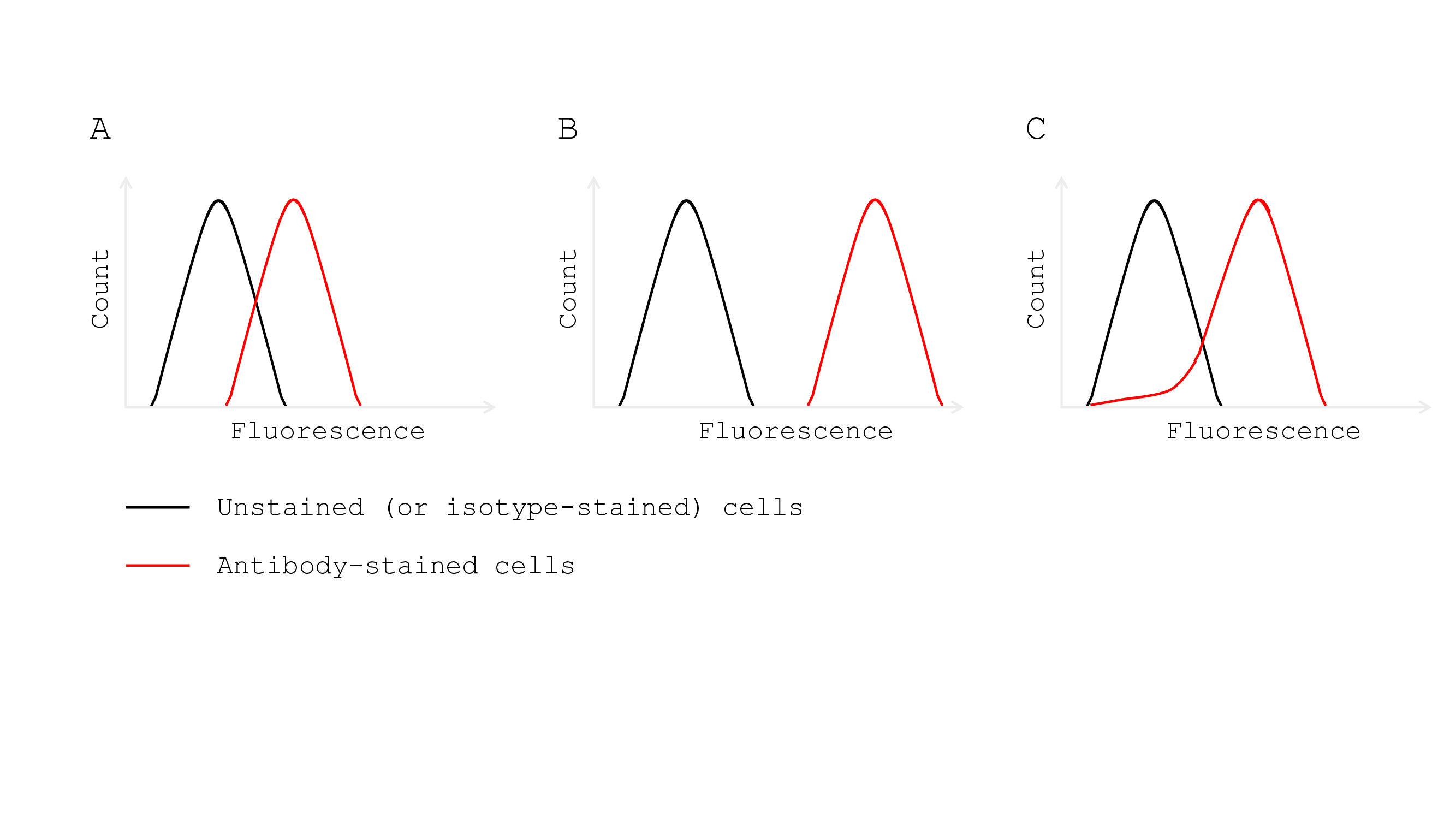}
        \captionsetup{font=small}
        \caption{
        \textbf{Flow cytometry task in HLE.}
        \textit{Q.} Three flow cytometry histograms (A, B, C) show cells
        stained with three different fluorescently-labelled antibodies.
        Which antibody stained \emph{all} cells in the sample?\quad
        \textit{A.} \textbf{A}
        }
    \end{subfigure}
    \captionsetup{font=small}
    \captionof{figure}{
    Example tasks from \projectname{} (\emph{left}) and HLE (\emph{right}) in the flow cytometry domain, contrasting a professional gating workflow with a more academic textbook/exam-like conceptual question. In immunology or cell therapy research, interpretation tasks like the one on the left help a scientist determine whether a treatment changed immune cell composition. This is direct interpretation of experimental output rather than a  conceptual test of principles like the right-hand task.
    }
    % \vspace*{-1em}
    \label{fig:vials-vs-hle-flow}
        \end{minipage}
\end{figure}

Existing benchmarks cover important aspects of scientific and multimodal visual reasoning, but largely target academic settings. MMMU and MMMU-Pro evaluate multimodal reasoning across many disciplines using academic materials such as exams and textbooks \citep{yue2024mmmu,yue2024mmmupro}. SciVQR \citep{scivqr} and SciVQA \cite{scivqa} source conceptual questions from scientific figures in textbooks, exams, academic competitions, and publications. More specialized benchmarks target particular biomedical visual settings: GMAI-MMBench evaluates VQA across medical imaging modalities and clinical tasks \citep{gmaimmb}, while MicroVQA focuses on  microscopy images \citep{microvqa}. These benchmarks provide important evaluations of scientific VQA, but do not target the visual artifacts and tasks encountered across professional life sciences work.

\FloatBarrier

Humanity's Last Exam (HLE) evaluates expert-level questions across many academic disciplines, including VQA tasks in Biology/Medicine and Chemistry  
\citep{phan2025lastexam}. However, an independent audit estimates that 29\% of HLE's biology and chemistry tasks have untrustworthy reference answers \citep{futurehouse2025hle}. In contrast, every reference answer in \projectname{} is rigorously vetted via careful reviews of multiple independent attempts at the task by different experts (\Cref{sec:qc}). 
Moreover, \Cref{fig:expert-assessment,fig:vials-vs-hle-flow} reveal that \projectname{} tasks are significantly more relevant to professional life sciences workflows than the tasks in HLE, which tend to be more academic/exam-like.
See more comparisons in \Cref{app:hle-comparison}.

\section{The \projectname{} Benchmark}
\label{sec:benchmark_description}

\projectname{} evaluates whether VLMs can interpret professional life science artifacts and use the visual evidence they contain to answer core scientific questions. Each task in the benchmark consists of one or more images, a question, and a short ground-truth answer (held-out and solely used for grading).

\subsection{Scoring model outputs}

\projectname{} is intended to evaluate VLMs with reasoning capabilities \citep{xu2025llava}. Our benchmark solely evaluates a model's generated final answer, not its intermediate reasoning process. 
To better reflect how surveyed scientists reported wanting to incorporate AI in their workflows, we design tasks to have open-domain text answers rather than multiple-choice.
Our benchmark uses semantic grading rather than exact-match evaluation because correct responses may differ from the reference answer in formatting, units, numerical representation, or wording while remaining scientifically equivalent (although questions are authored to minimize ambiguity where applicable).

As the ground-truth answers are short-form, we can effectively employ a simple LLM-as-a-judge approach to determine whether each model-generated response is correct  \citep{phan2025lastexam,simpleqa,browsecomp}. Using the grading prompt in \Cref{app:evaluation_details}, our LLM judge takes in the question, the model response, and the reference ground-truth answer. For certain tasks that experts determine have multiple acceptable answers, the LLM judge additionally receives this (expert-determined) acceptable range.  
Because the judging task is straightforward when given the short-form reference answer, we employ a fast, cost-effective model as the judge: OpenAI's GPT-5-mini.
We validate the LLM judge against independent human expert grading of 4{,}685 model-responses across all 161 tasks, where it agrees with human judgments on 99.9\% of cases. When testing various judge models for \projectname{} tasks, we observed little evidence of self-preference, verbosity, or style bias in the LLM-judge \citep{ye2025justice}.

\subsection{Domains covered in the benchmark}
\projectname{} covers a range of visual artifacts used across professional life sciences research.  Artifact domains were selected according to their criticality in high-value research workflows across the life sciences industry, particularly in the biotechnology and pharmaceutical sectors. For instance, many activities in pharmaceutical research depend on interpreting such artifacts, such as experimental characterization, assay development/screening, and drug discovery/development \citep{chen2024advancing,busby2020advancements,sinha2018drugdiscovery}.  

 \Cref{tab:domains} summarizes the artifact categories, representative image types, and the scientific fields in which they are commonly used.  These include phylogenetic trees, flow cytometry plots, blots and gels, plasmid maps, cell-counting and quantification images, protein-structure and binding views, and small-molecule structures. 
Each task is assigned a primary artifact domain and one or more secondary tags describing the specific artifact subtype, such as \textbf{Western blot} versus \textbf{SDS-PAGE} within the \textbf{blotting} domain, or \textbf{linear} versus \textbf{circular trees} within the \textbf{phylogenetics} domain.

\begin{table}[t]
\centering
\footnotesize
\renewcommand{\arraystretch}{1.15}
\setlength{\tabcolsep}{6pt}
\begin{tabularx}{\textwidth}{@{} >{\raggedright\arraybackslash}p{2.6cm} X X r @{}}
\toprule
\textbf{Domain} & \textbf{Representative image types} & \textbf{Scientific fields} & \textbf{Share} \\
\midrule
Phylogenetics       & Rectangular and circular phylogenetic trees; branch-length and clade diagrams; taxon and clade labels & Evolutionary biology, microbiology, genomics & 14.3\% \\
Flow cytometry      & Gating hierarchies, density/contour plots, FMO overlays, histograms, and multi-parameter panels & Immunology, biomedical research, cell biology & 14.3\% \\
Blotting            & Western blots, SDS-PAGE, and 2D gels; lane- and band-level intensity comparisons & Protein biochemistry, molecular biology & 14.3\% \\
Cell counting/Quantification       & Fluorescence images, colony and plaque images, blood smears, and confluence estimation & Molecular biology, microbiology, cell biology & 14.3\% \\
Plasmid maps        & Circular and linear plasmid maps; restriction maps; feature annotations and positions & Molecular biology, synthetic biology, virology & 13.0\% \\
Protein structure   & Protein--ligand binding diagrams, active-site views, 3D structure and residue contact maps & Structural biology, biochemistry, drug discovery & 13.0\% \\
Molecular structure & Small-molecule skeletal structures; substituent identification and reaction transformations & Medicinal chemistry, synthetic chemistry & 12.4\% \\
Other               & Pedigrees, Lateral flow assays, etc. & Genetics, systems biology, and related fields & 4.3\% \\
\bottomrule
\end{tabularx}
\captionsetup{font=small}
\caption{Artifact domains covered by \projectname{}, their representative image types, the scientific fields in which those artifacts are commonly found, and each domain's prevalence in the final \datasetsize{}-task benchmark.}
\label{tab:domains}
\vspace{-1em}
\end{table}

\subsection{Artifact sourcing and task construction}
\label{sec:artifact}

\projectname{} includes artifacts that are unpublished real experimental data, artifacts created through procedural generation with scientific software, and artifacts sourced from open-access publications with a CC-BY license. 
The benchmark emphasizes messy data artifacts that are typical in real-world research.
Procedural generation using scientific software was used for select artifact types: phylogenetic trees, plasmid maps, and blots (all of which were expert-verified as realistic). 

To create tasks for the benchmark, domain experts provide one or more images and a question that reflect their actual work experience, and then provide the ground truth answer for the question. 
Beyond representing core tasks from their work, the questions are designed to be answerable from the provided artifact while requiring both the extraction of relevant visual evidence and the application of scientific knowledge or reasoning.

\subsection{Task contributors} 

\projectname{} was developed by 31 contributors\footnote{Benchmark contributors were sourced from the Handshake talent network: \url{https://joinhandshake.com/ai}} with %broad collective expertise across the life sciences, each possessing 
deep professional expertise in at least one of the category domains. 
Contributors trained at institutions including Harvard Medical School, MIT, Stanford, UC Berkeley, UCLA, NYU, Yale, and Oxford, and had prior work experience at organizations including Johnson \& Johnson, Amgen, Bristol Myers Squibb, Abbott Laboratories, Mayo Clinic, the National Institutes of Health, and the U.S.\ Food and Drug Administration. Overall, 87\% had doctoral-level training, and 74\% had more than six years of professional work experience as a researcher in the life sciences industry. All contributors passed a rigorous qualification assessment, demonstrating scientific expertise in at least one of the following areas: immunology, microbiology, molecular biology, genetics and genomics, biochemistry, oncology, virology, neuroscience, bioinformatics, or drug discovery. Contributors created and reviewed benchmark tasks only within their areas of expertise, ensuring that the tasks reflect critical activities from their own professional experience.

\subsection{Quality control pipeline}
\label{sec:qc}

Each task passes through multiple stages of human review to ensure it is realistic and correctly specified \citep{futurehouse2025hle}. A domain expert constructs the task by selecting or generating the relevant artifact, writing the question, and providing a reference answer. A second (and often third) expert with relevant domain expertise then answers the same task independently, without seeing the original answer.
These independent task attempts (along with writing detailed justifications for their answers) took experts 16 minutes on average. 
Another expert reviewer checks that the images are understandable and realistic, and the question represents a meaningful life sciences task that is well-specified and answerable from the provided information. In addition, reviewers grade the correctness of model-generated answers and the independent expert attempts, to finalize the ground-truth answer and whether a range of values should be considered acceptable. 
A task is retained only if the independent experts' answers agree with the original reference answer or experts reach a consensus in follow-up reviews.  

During benchmark construction, we additionally ensured task diversity by checking whether incoming tasks are similar to others in the benchmark. We also ensure tasks are not already found online (i.e.\ in frontier models' pretraining data). 
Whenever the prevalence of tasks from certain domains fell below their target representation (estimated via expert surveys of common life sciences workflows), we encouraged the relevant experts to create more tasks in these underrepresented domains until the benchmark reflected the target domain distribution in \Cref{tab:domains}.

\subsection{Expert assessment of professional relevance}
\label{sec:expert-assessment}

Beyond validating task correctness, we also asked expert reviewers to rate the professional relevance of \projectname{} tasks within their domain of expertise (rating tasks created by other experts, which they were not familiar with). 
We asked these same experts to similarly assess 118 VQA tasks from HLE's Biology/Medicine and Chemistry subset\footnote{No reviewer felt familiar with the 119th such HLE task, so our expert assessment of professional relevance omitted it.}. Reviews were conducted blind, such that reviewers were unaware of the source of a task while rating it.

\Cref{fig:expert-assessment} and \Cref{app:expert-survey} detail the survey questions and results, revealing that experts are more likely to classify \projectname{} artifacts as resembling professional work than those in HLE  (\professionalshare{} versus \hleprofessionalshare{}). Experts also reported encountering \projectname{} artifact types more frequently in their work, with \projectnameweeklydailyshare{} selecting weekly or daily compared with \hleweeklydailyshare{} for HLE.  For \commercialshare{} of \projectname{} tasks, experts agreed that interpreting the artifact has commercial or economic value, compared with \hlecommercialshare{} for HLE. These assessments indicate that visual tasks in  \projectname{} are not only scientifically meaningful, but also closely connected to professional life science practice.

\section{Evaluation Setup}
\label{sec:experiments}

We evaluate several frontier VLMs from different providers. 
All multimodal models receive the same prompt, which includes only the image(s) and the question. \Cref{app:evaluation_details} provides the full prompt and configurations used.  
We independently run each model three times on each task, and all results report its average accuracy across the three rollouts.

\section{Results}
\label{sec:results}
\subsection{Overall model performance}
\begin{figure*}[t]
    \vspace{-1em}
    \centering
    \includegraphics[width=\textwidth]{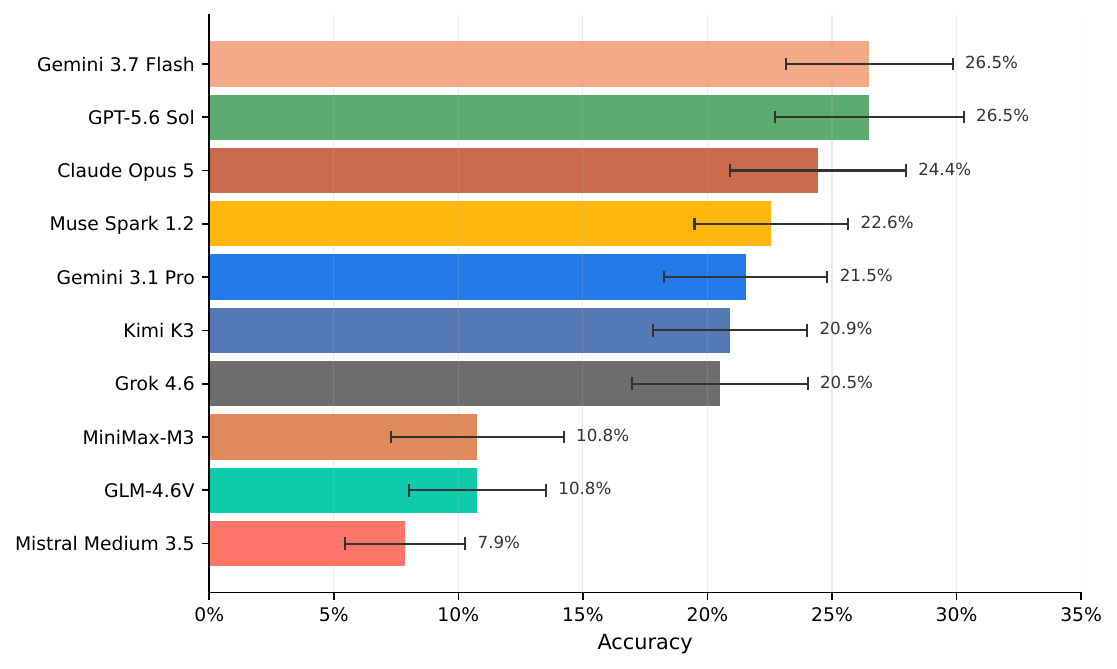}
    \captionsetup{font=small}
    \vspace{-1em}
    \caption{Performance of various vision-language models on \projectname{} (averaged across 3 independent rollouts per task). Error bars show 95\% confidence intervals computed by bootstrap resampling across the three rollouts for each task (reflecting variability in \emph{accuracy} under re-evaluation with one attempt per task). Claude Fable 5 could not be evaluated as its safety guardrails blocked most of the requests from this benchmark.}
    \label{fig:leaderboard}
    \vspace{-1em}
\end{figure*}

\Cref{fig:leaderboard} shows overall accuracy of each vision-language model on \projectname{}. The top-performing models, \bestmodel{}, achieve only \bestscore{}, with Gemini 3.7 Flash incurring lower costs (\Cref{fig:cost-tokens}). The Pass\textasciicircum{}3 results in \Cref{fig:pass3} show that these models are only able to handle under 17\% of \projectname{} tasks if we require that all 3 rollouts from the model are correct, a basic requirement for trusting a model in critical life sciences research.

Performance differs substantially across domains (\Cref{tab:per-domain}). Phylogenetics is the strongest domain for the leading models, with GPT-5.6 Sol and Muse Spark 1.2 reaching 43.5\% and 34.8\% accuracy, respectively. Results are lower in most other domains. The best scores are 38.1\% on plasmid maps (Muse Spark 1.2), 33.3\% on both protein structure (Muse Spark 1.2) and flow cytometry (Kimi K3 and Grok 4.6, tied), 31.7\% on molecular structure (GPT-5.6 Sol), 30.4\% on blotting (GPT-5.6 Sol and Claude Opus 5), and 21.7\% on cell counting/quantification (Gemini 3.7 Flash). Current VLMs generally struggle the most in cell counting/quantification tasks and protein structure tasks.

\begin{table*}[t]
\centering
\footnotesize
\renewcommand{\arraystretch}{1.2}
\setlength{\tabcolsep}{4pt}
\scriptsize
\begin{tabularx}{\textwidth}{@{} l @{\hspace{2pt}} *{8}{>{\centering\arraybackslash}X} @{}}
\toprule
\textbf{Model} &
\textbf{Blotting} &
\shortstack{\textbf{Cell Count.}\\\textbf{/ Quant.}} &
\shortstack{\textbf{Flow}\\\textbf{Cytometry}} &
\shortstack{\textbf{Molecular}\\\textbf{Structure}} &
\textbf{Phylogenetics} &
\shortstack{\textbf{Plasmid}\\\textbf{Maps}} &
\shortstack{\textbf{Protein}\\\textbf{Structure}} &
\textbf{Other} \\
\midrule
GPT-5.6 Sol         & \textbf{30.4\%} & 15.9\%          & 29.0\%          & \textbf{31.7\%} & \textbf{43.5\%} & 22.2\%          & 15.9\%          & \textbf{14.3\%} \\
Gemini 3.7 Flash    & 29.0\%          & \textbf{21.7\%} & 31.9\%          & 30.0\%          & 29.0\%          & 23.8\%          & 23.8\%          & \textbf{14.3\%} \\
Claude Opus 5       & \textbf{30.4\%} & 20.3\%          & 27.5\%          & 18.3\%          & 31.9\%          & 25.4\%          & 19.0\%          & \textbf{14.3\%} \\
Muse Spark 1.2      & 15.9\%          &  2.9\%          & 21.7\%          & 15.0\%          & 34.8\%          & \textbf{38.1\%} & \textbf{33.3\%} & \textbf{14.3\%} \\
Gemini 3.1 Pro      & 24.6\%          & 18.8\%          & 14.5\%          & 20.0\%          & 31.9\%          & 25.4\%          & 17.5\%          & \textbf{14.3\%} \\
Kimi K3             & 21.7\%          & 10.1\%          & \textbf{33.3\%} & 23.3\%          & 27.5\%          & 19.0\%          & 12.7\%          & \textbf{14.3\%} \\
Grok 4.6            & 14.5\%          & 15.9\%          & \textbf{33.3\%} & 13.3\%          & 30.4\%          & 19.0\%          & 17.5\%          & \textbf{14.3\%} \\
GLM-4.6V            & 18.8\%          & 11.6\%          & 10.1\%          &  3.3\%          & 13.0\%          & 12.7\%          &  6.3\%          &  4.8\%          \\
MiniMax-M3          &  8.7\%          &  8.7\%          & 13.0\%          &  5.0\%          & 17.4\%          & 15.9\%          &  7.9\%          &  4.8\%          \\
Mistral Medium 3.5  & 15.9\%          &  4.3\%          &  8.7\%          &  8.3\%          &  7.2\%          &  7.9\%          &  4.8\%          &  0.0\%          \\
\bottomrule
\end{tabularx}
\captionsetup{font=small}
\caption{Per-domain accuracy on \projectname{} achieved by each vision-language model. Bold indicates the best performance in each domain.}
\label{tab:per-domain}
\end{table*}

\begin{table}[t]
\centering
\scriptsize
\renewcommand{\arraystretch}{1.2}
\setlength{\tabcolsep}{3pt}
\begin{tabularx}{\textwidth}{@{} l *{7}{>{\centering\arraybackslash}X} @{}}
\toprule
\textbf{Model} & \textbf{Visual quantification error} & \textbf{Incorrect value or feature selection} & \textbf{Overlooked evidence} & \textbf{Representation misinterpretation} & \textbf{Quantitative reasoning error} & \textbf{Misapplied principle} & \textbf{Fabricated finding} \\
\midrule
GPT-5.6 Sol          & 41.8\% & 14.2\% & 15.6\% & 14.2\% &  6.4\% &  3.5\% &  4.3\% \\
Gemini 3.7 Flash     & 35.3\% & 18.0\% & 18.0\% & 16.5\% &  8.3\% &  3.0\% &  0.8\% \\
Claude Opus 5        & 36.8\% & 16.9\% & 17.6\% & 14.0\% & 10.3\% &  4.4\% &  0.0\% \\
Muse Spark 1.2       & 38.3\% & 26.3\% & 12.0\% & 16.5\% &  3.8\% &  2.3\% &  0.8\% \\
Gemini 3.1 Pro       & 34.3\% & 12.9\% & 20.7\% & 15.0\% & 10.7\% &  5.0\% &  1.4\% \\
Kimi K3              & 33.1\% & 20.9\% & 15.8\% & 16.5\% &  7.9\% &  4.3\% &  1.4\% \\
Grok 4.6             & 33.1\% & 17.2\% & 19.3\% & 17.9\% &  7.6\% &  4.8\% &  0.0\% \\
GLM-4.6V             & 28.3\% & 19.7\% & 17.8\% & 17.8\% &  9.9\% &  5.9\% &  0.7\% \\
MiniMax-M3           & 32.1\% & 18.6\% & 19.2\% & 20.5\% &  5.8\% &  3.2\% &  0.6\% \\
Mistral Medium 3.5   & 28.2\% & 19.9\% & 15.4\% & 21.8\% &  7.7\% &  6.4\% &  0.6\% \\
\bottomrule
\end{tabularx}

\captionsetup{font=small}
\caption{Distribution of failure modes by model.}
\label{tab:failure-modes}
\vspace{-1em}
\end{table}

\begin{figure*}[!b]
  \centering

  % ---------------- row 1 ----------------
  \begin{subfigure}[t]{0.48\textwidth}
    \centering
    \includegraphics[width=\linewidth]{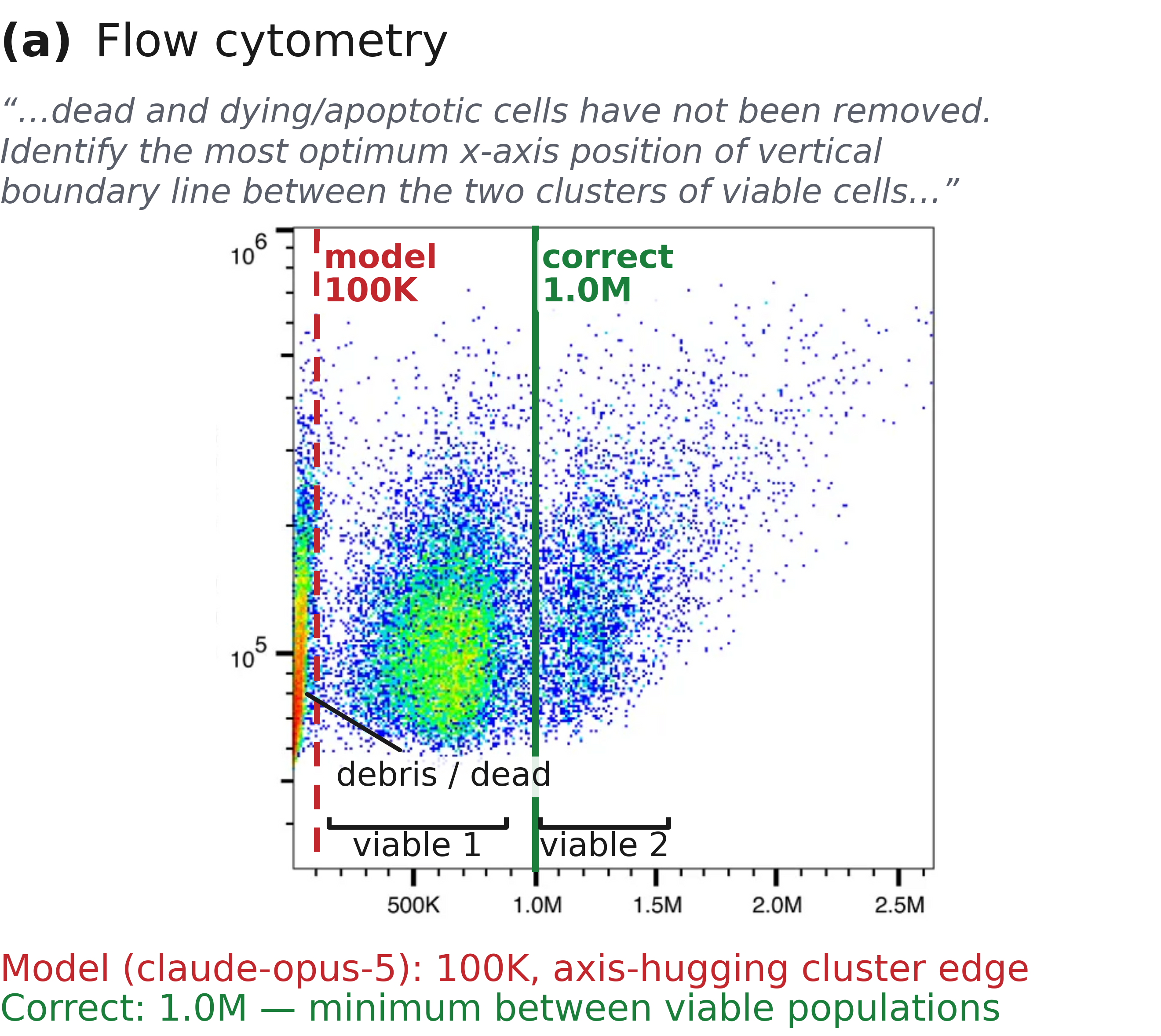}
    \captionsetup{font=small}
    \caption{\textbf{Wrong population.} The model selects the debris--viable
    boundary instead of the boundary between the two viable populations.}
    \label{fig:fail-flow}
  \end{subfigure}
  \hfill
  \begin{subfigure}[t]{0.48\textwidth}
    \centering
    \includegraphics[width=\linewidth]{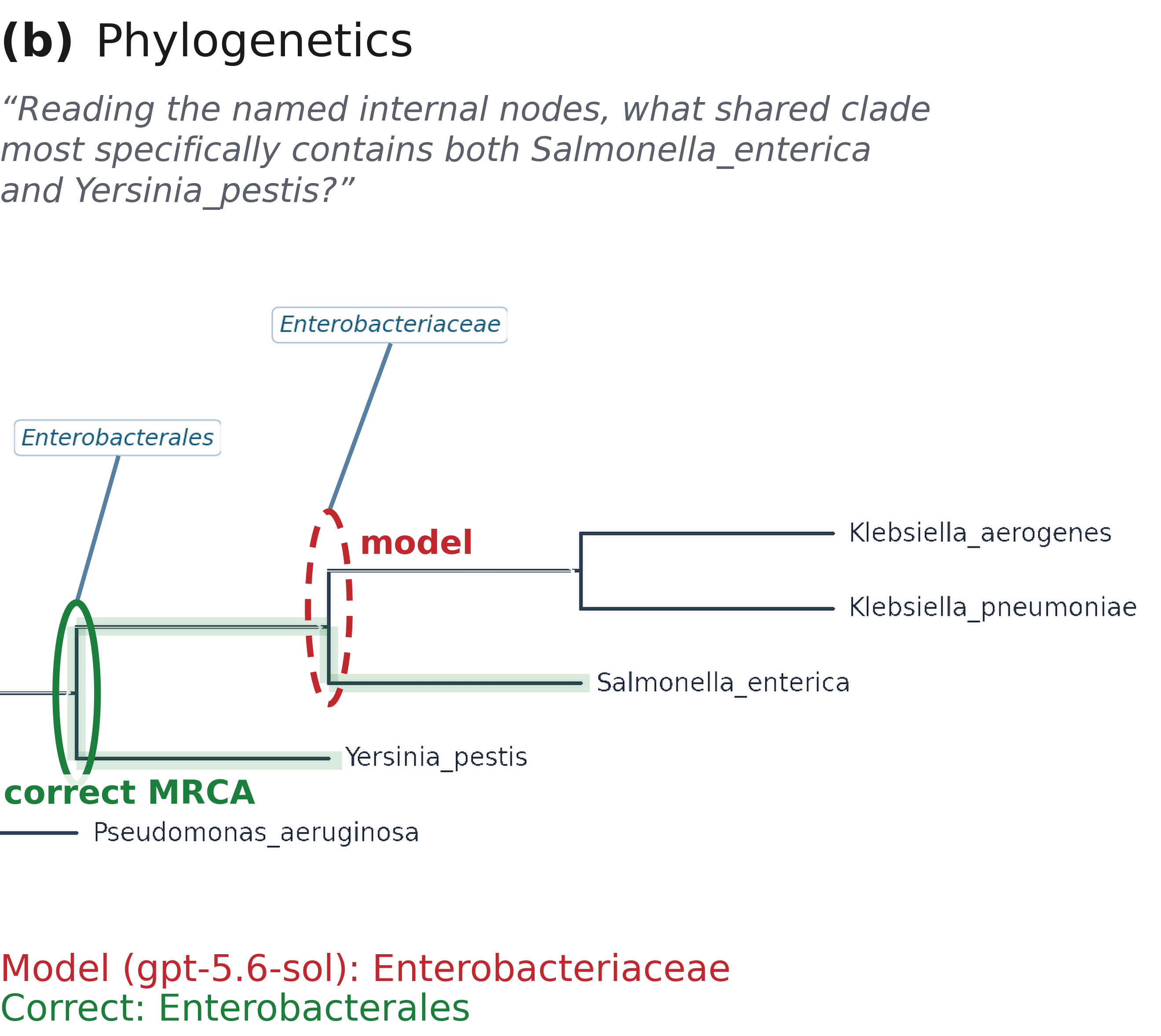}
    \captionsetup{font=small}
    \caption{\textbf{Topology error.} The model relies on spatial proximity
    rather than the tree topology.}
    \label{fig:fail-phylo}
  \end{subfigure}

  \vspace{0.6em}

  % ---------------- row 2 ----------------
  \begin{subfigure}[t]{0.48\textwidth}
    \centering
    \includegraphics[width=\linewidth]{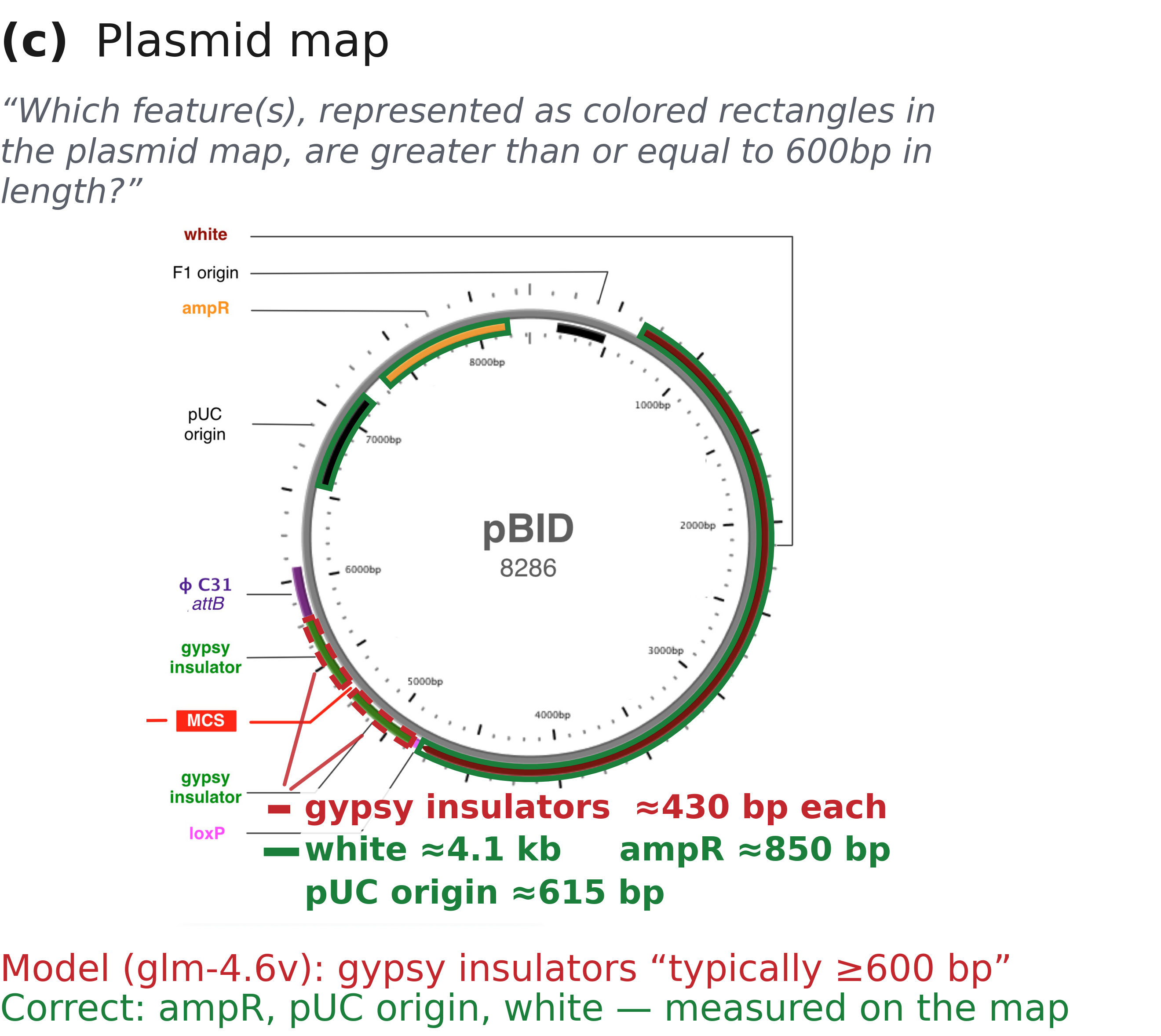}
    \captionsetup{font=small}
    \caption{\textbf{Prior over evidence.} The model uses a typical feature
    length rather than measurements from the plasmid map.}
    \label{fig:fail-plasmid}
  \end{subfigure}
  \hfill
  \begin{subfigure}[t]{0.48\textwidth}
    \centering
    \includegraphics[width=\linewidth]{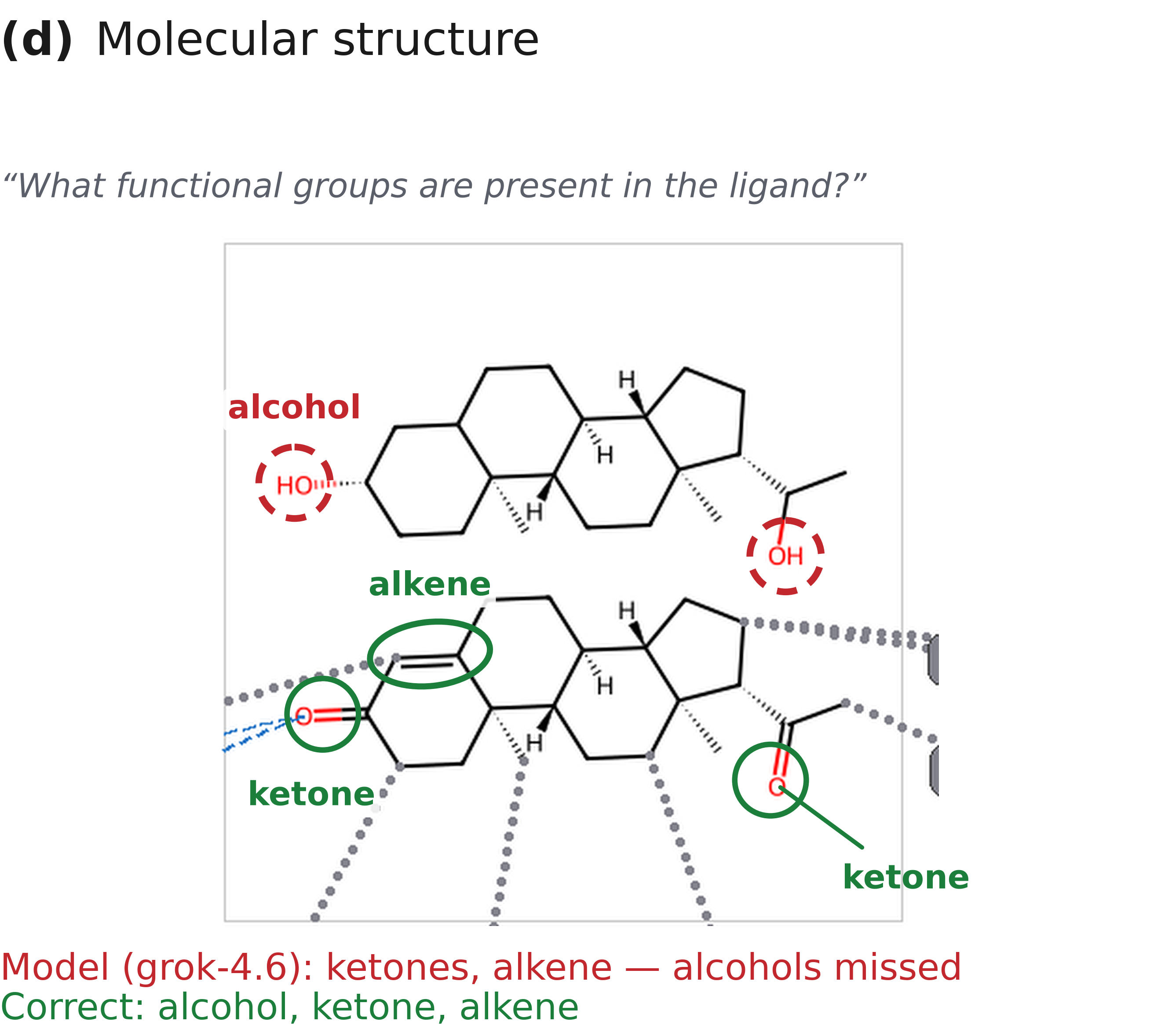}
    \captionsetup{font=small}
    \caption{\textbf{Feature not perceived.} The model correctly names the two ketone carbonyls and the alkene, but asserts that no alcohol groups are present on the ligand, missing the two hydroxyls on the steroid A-ring and side chain.}
    \label{fig:fail-mol}
  \end{subfigure}
    \captionsetup{font=small}
  \caption{Representative model failures beyond visual extraction. Models can
  identify relevant features in scientific artifacts yet still use them
  incorrectly in downstream reasoning. Each panel shows the model's reading
  in red, and the correct reading supported by the artifact in green.}
  \label{fig:failure-examples}
\end{figure*}

\subsection{Failure mode analysis}
\label{sec:error-analysis}

\Cref{fig:failure-examples} shows that failures are not limited to extracting information from the artifact. Models may identify salient visual evidence yet still apply the wrong representational convention, substitute prior expectations for evidence in the artifact, or fail to use the extracted information consistently in the final inference. These examples highlight a distinct challenge in scientific visual reasoning: correctly interpreting visual evidence in the context of the underlying scientific representation.

To understand where models fail, we assign a primary failure mode for each model on each task with at least one incorrect rollout via axial coding with an LLM classifier (see \Cref{app:failuredetails} for details). We consider seven failure categories: \emph{Visual quantification error}, for errors in counting objects such as cells, colonies, or bands; \emph{Incorrect value or feature selection}, for reading the wrong value, label, lane, gate, or feature; \emph{Overlooked evidence}, for missing relevant information visible in the artifact; \emph{Representation misinterpretation}, for incorrectly reading the structure or conventions of a plot, tree, plasmid map, or other scientific representation; \emph{Quantitative reasoning error}, for arithmetic, bound, or unit errors after the relevant values have been identified; \emph{Misapplied scientific principle}, for applying an incorrect scientific rule or technique; and \emph{Fabricated finding}, for introducing an element or measurement not supported by the artifact.

Most failures occur while models are reading or interpreting the artifact. Across models, 83--93\% of errors fall into the first four categories in \Cref{tab:failure-modes}. Visual quantification is the most common failure mode for every model, accounting for 28.2--41.8\% of errors. Incorrect value or feature selection, overlooked evidence, and representation misinterpretation are also common across models.

Quantitative reasoning errors account for 3.8--10.7\% of failures, while misapplied scientific principles account for 2.3--6.4\%. Fabricated findings are rare, accounting for at most 4.3\% of errors. Overall, the failure analysis suggests that the main difficulty is often identifying and correctly interpreting the information encoded in the artifact, rather than carrying out the subsequent calculation or applying a scientific principle.

\subsection{Evaluating tool-assisted agents}
\label{sec:agent}

Our VQA benchmark focuses on evaluating VLMs, with tasks being completed via single-turn inference calls to multimodal reasoning models. Human scientists can handle these tasks relatively quickly by simply viewing the image without relying on software or external tools. Competent scientific AI models should be able to do the same.
Modern life sciences research is starting to generate visual artifacts at scale: high-content screening campaigns such as Cell Painting routinely produce images across hundreds of thousands of chemical and genetic perturbations \citep{bray2016cell, chandrasekaran2023jump}, and interpretation of assay readouts is a recurring, throughput-limiting step in drug discovery pipelines \citep{busby2020advancements}. A model that interprets these artifacts with expert-level accuracy and machine cost/speed can remove a bottleneck that is currently dependent on human experts. A less efficient AI system will yield less acceleration of the experimental loop.

Nonetheless, our previous failure-mode analysis raises the question: how many errors made under direct visual evaluation can be recovered when models are given additional ways to iteratively inspect the same artifact? To answer this, we evaluate the same multimodal models in an \emph{agentic setup with code execution}, where models can loop and iteratively crop, zoom, transform, measure, and otherwise programmatically inspect the supplied image before answering.

We evaluate several model and agent-harness configurations including OpenAI's Codex\footnote{\url{https://openai.com/codex/}}, Claude Code\footnote{\url{https://claude.com/product/claude-code}}, and OpenCode\footnote{\url{https://github.com/anomalyco/opencode}} (see \Cref{app:tool-assisted} for details). \Cref{tab:tool-assisted} shows that agentic tool access markedly improves performance for most models. 
However, these gains come with substantially greater token usage and costs: tool-assisted runs use 10--432$\times$ more tokens per attempt than direct evaluation, depending on the model--harness configuration. GLM-4.6V is an exception: its overall performance decreases with tool access. 

These results suggest that many failures of the reasoning VLM are recoverable with additional inference-time resources and access to software tools. However, these gains incur a significant associated computational cost burden, despite the fact that scientists can quickly interpret the underlying artifacts in the relevant domains without such tools. For today's frontier VLMs, this indicates that there remains significant room for improvement in the base model's raw visual localization and reasoning capabilities, particularly as it pertains to such artifacts.

\begin{table}[t]
\centering
\footnotesize
\setlength{\tabcolsep}{3.5pt}
\begin{tabular}{llcccccc}
\toprule
\textbf{Model} &
\textbf{Harness} &
\textbf{Direct VLM} &
\textbf{Tool-Assisted} &
$\mathbf{\Delta}$ &
\textbf{Tok$_{\mathrm{D}}$} &
\textbf{Tok$_{\mathrm{T}}$} &
\textbf{Tok$\times$} \\
\midrule
GPT-5.6 Sol
    & Codex       & 26.5 & 49.1 & +22.6 & 4.1k & 165k &  40$\times$ \\
GPT-5.6 Sol
    & OpenCode    & 26.5 & 40.4 & +13.9 & 4.1k &  47k &  11$\times$ \\
\midrule
Claude Opus 5
    & Claude Code & 24.4 & 64.2 & +39.8 & 5.3k & 959k & 182$\times$ \\
Claude Opus 5
    & OpenCode    & 24.4 & 65.8 & +41.4 & 5.3k & 737k & 139$\times$ \\
\midrule
Gemini 3.1 Pro
    & OpenCode    & 21.5 & 35.2 & +13.7 & 3.5k & 1.0M & 285$\times$ \\
Gemini 3.7 Flash
    & OpenCode    & 26.5 & 60.7 & +34.2 & 3.7k & 1.6M & 432$\times$ \\
Grok 4.6
    & OpenCode    & 20.5 & 45.8 & +25.3 & 10.4k & 382k &  37$\times$ \\
Kimi K3
    & OpenCode    & 20.9 & 64.2 & +43.3 & 20.2k & 557k & 28$\times$ \\
GLM-4.6V
    & OpenCode    & 10.8 &  7.0 & $-$3.7 & 5.5k &  56k &  10$\times$ \\
\bottomrule
\end{tabular}
\captionsetup{font=small}
\caption{
Performance of tool-assisted agents on \projectname{}, 
 across seven models run in different harnesses.
\textbf{Direct VLM} vs.\ \textbf{Tool-Assisted} report accuracy (\%) under the single/direct VLM call vs.\ agentic tool-assisted settings, respectively. Values are the average over $K{=}3$
rollouts. 
$\mathbf{\Delta}$ reports the absolute change in accuracy in percentage points.
\textbf{Tok$_{\mathrm{D}}$} and \textbf{Tok$_{\mathrm{T}}$} report mean tokens
per attempt in the direct-VLM and tool-assisted settings, respectively;
\textbf{Tok$\times$} reports the ratio of tool-assisted to direct-VLM token usage.
}
\label{tab:tool-assisted}
\end{table}

\section{Conclusion}
\label{sec:conclusion}

This paper introduces \projectname{}, a benchmark for evaluating how accurately VLMs can interpret visual artifacts from professional life sciences workflows that carry direct commercial and experimental consequences. 
% Its \datasetsize{} tasks span diverse artifact types and require models to identify relevant visual evidence and apply scientific reasoning. The tasks are constructed by domain experts, independently validated for clarity and answer consistency, and assessed by PhD researchers and industry scientists for professional relevance.
Across leading available multimodal models, accuracy remains limited. Our error analysis indicates that many failures arise before higher-level reasoning: models often identify the general artifact type but misread labels, values, regions, or relationships needed to answer the question correctly. These results suggest that reliable interpretation of scientific images remains a substantial bottleneck for deploying VLMs in life sciences workflows.

Our agentic tool-assisted evaluation further decomposes the deficit of current frontier models into two gaps. First, a \emph{perception gap}: when the same models are given code-execution tools to iteratively crop, zoom, and measure the artifact, accuracy improves by up to 43 points, indicating that much of the required scientific knowledge is present but cannot be applied through the neural network's direct visual inference \citep{tong2024eyes}. However, these gains come at $10$--$432\times$ the token cost of direct inference, for interpretations that trained scientists can resolve in minutes. Second, a residual \emph{interpretation gap}: even with unrestricted iterative inspection and programmatic tool use, the best agent solves only 65\% of tasks, with failures concentrated in reading scientific representations and conventions along with proper application of other domain knowledge. Because every VIALS task is grounded
in a real workflow with an expert-vetted answer, the benchmark measures both gaps against the standard that professional work requires.

\textbf{\projectname{} has several limitations}. The benchmark is limited in scale and does not capture the full range of visual artifacts or interpretation activities that encompass all life sciences research, focusing instead on high-frequency tasks in certain high-value workflows. The benchmark focuses on tasks which are relatively straightforward for scientists with the relevant Ph.D.\ to quickly accomplish, whereas professional research involves many other scientific artifact interpretations that are highly nontrivial due to ambiguity and experiment noise. \projectname{} also does not assess how well models can interpret scientific figures from publications nor academic textbooks/exams, which can require significantly more complex reasoning for humans to do accurately. Upon interpreting the visual artifacts from a life sciences workflow, human scientists must subsequently decide how to adjust their research plan---an important capability to evaluate models for that we leave to future benchmarks.

We hope \projectname{} advances the development of models that can interpret scientific artifacts accurately enough to be useful in professional life sciences research. To expand the benchmark, we are privately developing hundreds of additional tasks that will enable future benchmark refreshes and prevent overfitting to the current public benchmark.

\section{Acknowledgments}
The authors gratefully acknowledge the contributions of Le Li, Brian Shing, Roberto Alers-Velazquez, Alexander Y. Yang, Scott Espich, Shayan Mohammadmoradi, Austin Fergusson, Vivek Kohar, Aida Javidan, Adit Naor, Ebun Maria Omole-Ohonsi, Joshua Choi, Jessica M. Novotny, Anh Nguyen, Jens Heller, Brendan Geoffrey John Todd, Carmel T. Chan, Elliot Akama-Garren, Zhuoran Zhang, Aashiya Kolengaden, Elisabeth Diatta-Holgate, Faiz Ur Rahman, and Yuri Sarma. We thank each of them for their valuable support, insight, and dedication throughout the course of this work.

\clearpage 
\bibliographystyle{abbrvnat}
\bibliography{references} % expects references.bib

% =====================================================================
% Appendix
% =====================================================================
\clearpage
\appendix

\renewcommand{\thefigure}{A\arabic{figure}}
\renewcommand{\thetable}{A\arabic{table}}
\setcounter{figure}{0}
\setcounter{table}{0}

\thispagestyle{empty}

\begin{center}
\huge \textbf{Appendix}
\end{center}
\FloatBarrier

% Make appendix Cref show Appendix instead of Section:
\crefalias{section}{appendix}
\crefalias{subsection}{appendix}
\crefalias{subsubsubsection}{appendix}

\section{Additional Results}
\label{app:additional-results}

\subsection{Pass@3 and Pass\textasciicircum{}3 performance, costs, and token efficiency}
\label{app:pass3}
The main leaderboard reports mean accuracy across three rollouts. We additionally report Pass@3, the fraction of tasks answered correctly in at least one of the three attempts, and Pass\textasciicircum{}3, the fraction answered correctly in all three attempts. 

\begin{figure}[H]
    \centering
    \captionsetup{font=small}
    \includegraphics[width=\linewidth]{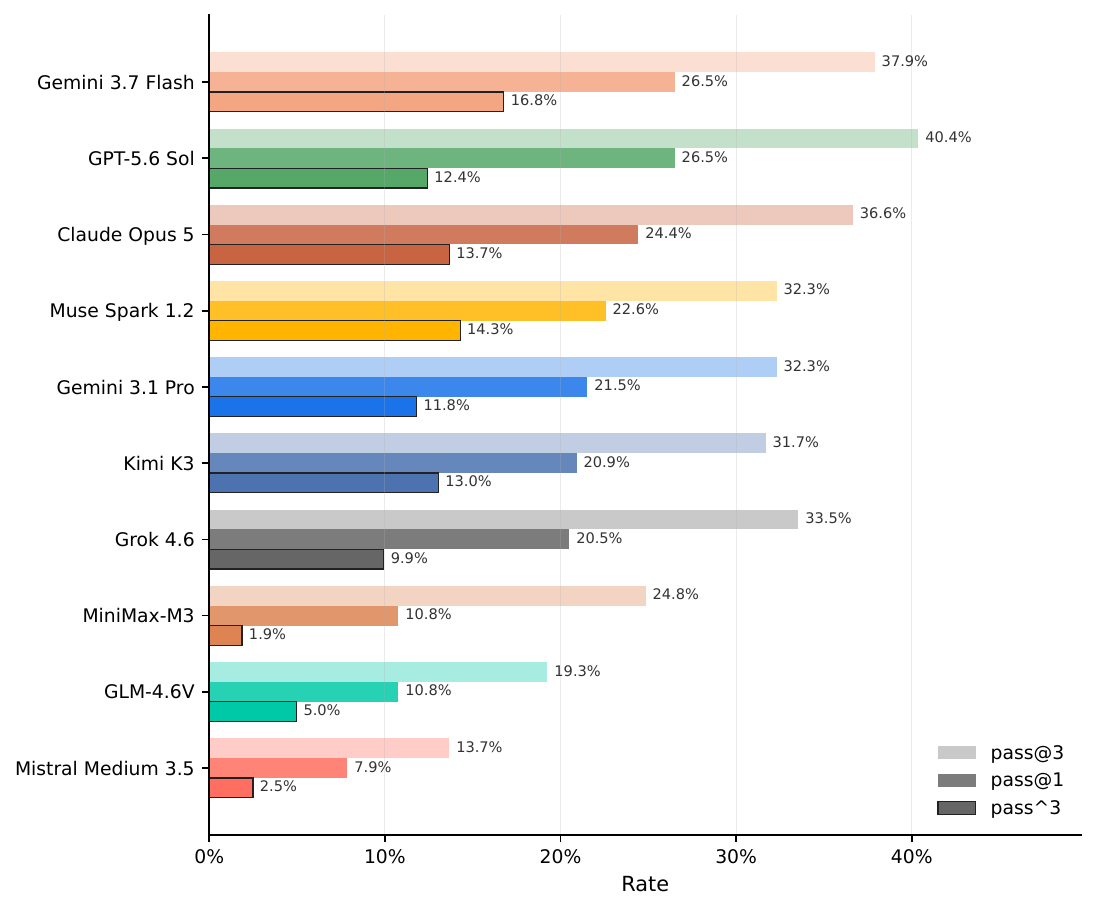}
    \caption{Mean accuracy (Pass@1), Pass@3, and Pass\textasciicircum{}3 achieved by different vision-language models. Pass@3 measures whether a task is solved correctly in at least one of three rollouts, while Pass\textasciicircum{}3 measures whether it is consistently solved across all three.}
    \label{fig:pass3}
\end{figure}

\Cref{fig:pass3} compares mean accuracy, Pass@3, and Pass\textasciicircum{}3. Across models, Pass@3 tops out around 40\%, while Pass\textasciicircum{}3 remains below 17\%. The large gap between these metrics shows that models often fail to reproduce correct answers across repeated attempts. This matters in professional workflows, where a model that is correct once but unreliable on the next run is difficult to depend on.

\begin{figure}[H]
    \centering
    \includegraphics[width=\linewidth]{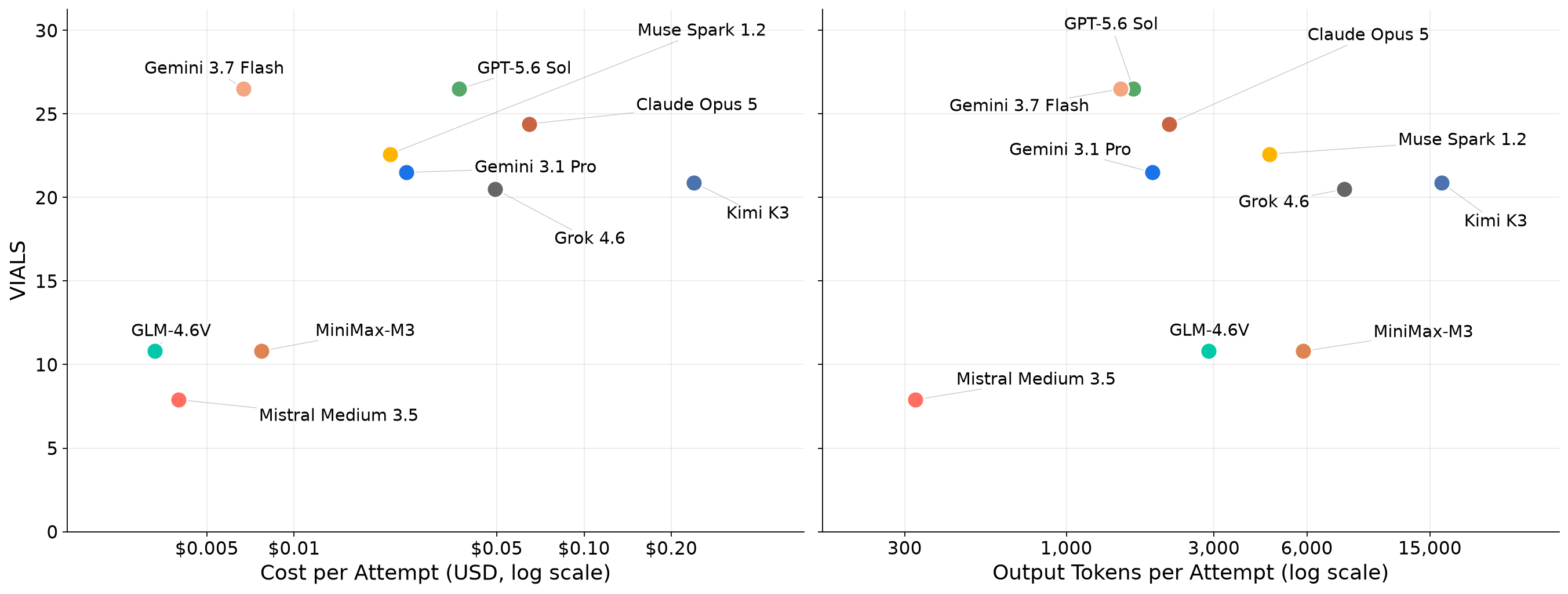}
    \captionsetup{font=small}
    \caption{\textbf{Cost and output-token efficiency on \projectname{}.}
Each point represents one model; accuracy is plotted against inference cost per attempt (\textbf{left}) and output tokens per attempt (\textbf{right}).}
    \label{fig:cost-tokens}
    \vspace{-1em}
\end{figure}

\Cref{fig:cost-tokens} compares model accuracy with inference cost and output-token usage. We observe substantial variation in efficiency across models, with higher cost or longer outputs not consistently corresponding to higher accuracy.

\subsection{Self-reported model confidence}
\label{app:confidence}

In high-stakes AI applications like life sciences research, it is valuable to have a calibrated measure of uncertainty in model responses to know when they can be trusted \citep{chen2024quantifying}. 
We evaluate whether the confidence reported by the model is predictive of the correctness of the response in \projectname{} \citep{tian2023just}.
For each task (and for all models), we augment the prompt in \Cref{app:evaluation_details} by additionally asking the model to verbalize a confidence score for its answer, expressed on a five-point Likert scale.

\begin{tcolorbox}[
    title={Addition to task prompt for self-reporting confidence},
    breakable,
    colback=gray!5,
    colframe=gray!40,
    boxrule=0.4pt,
    colbacktitle=gray!10,
    coltitle=black,
    fonttitle=\bfseries\small,
    fontupper=\small\rmfamily
]

% {\small\bfseries User message\par}
% \smallskip

After providing your final answer, report your confidence that the final
answer is correct on the following scale:

\smallskip

\noindent
1 = Very low: essentially guessing; substantial uncertainty

\noindent
2 = Low: significant uncertainty; answer may well be incorrect

\noindent
3 = Moderate: plausible answer, but meaningful ambiguity remains

\noindent
4 = High: likely correct; only limited uncertainty remains

\noindent
5 = Very high: clear interpretation; little meaningful uncertainty

\smallskip

\noindent
Format your response exactly as:

\smallskip

\noindent
\texttt{REASONING: <your reasoning>}

\noindent
\texttt{ANSWER: <final answer>}

\noindent
\texttt{CONFIDENCE: <1--5>}

\end{tcolorbox}

Before grading, we strip the \texttt{CONFIDENCE:} line from the candidate answer and apply the same grading procedure as in the direct evaluation. We separately extract the integer following the final \texttt{CONFIDENCE:} marker for the confidence analysis. Responses with missing, non-integer, or out-of-range confidence values are excluded. We compare accuracy between low- or moderate-confidence responses ($c \leq 3$) and high-confidence responses ($c \geq 4$).

\begin{figure}[H]
    \centering
    \includegraphics[width=\linewidth]{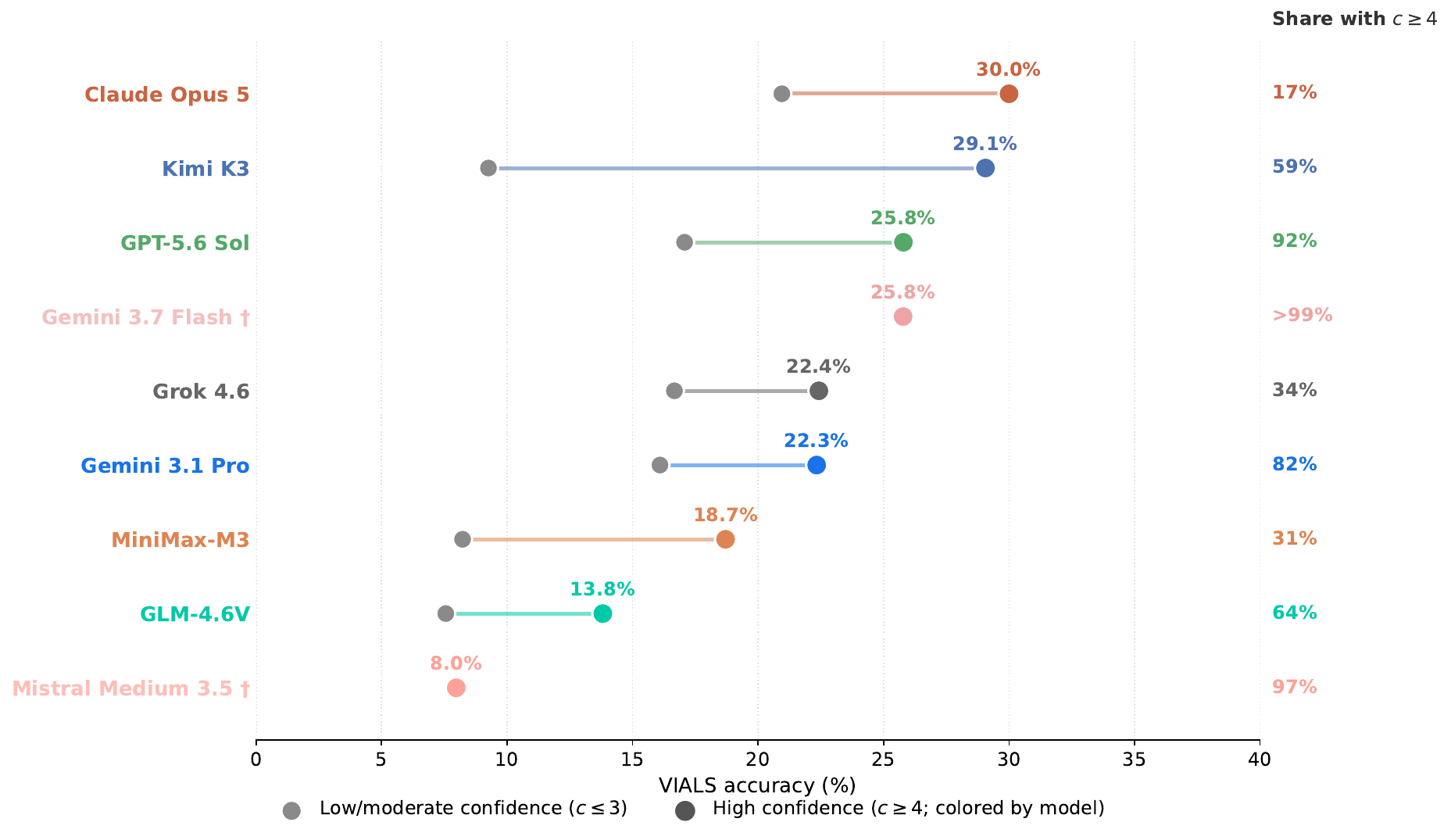}
    \captionsetup{font=small}
    \caption{
    Accuracy of model responses amongst those with high vs.\ low self-reported confidence. Gray points show the 
    accuracy of responses with low confidence ($c \leq 3 / 5$), while colored points show the
    accuracy of responses with high confidence ($c \geq 4 / 5$). The right column reports the share of
    model responses that were self-reported as high-confidence ($c \geq 4 / 5$). Models marked with $\dagger$ reported too
    few low-confidence responses for a reliable comparison.
    }
    \label{fig:confidence}
\end{figure}

Of the nine models shown in \Cref{fig:confidence}, seven produce enough lower-confidence responses for a meaningful comparison. For these models, high-confidence responses ($c \geq 4$) are about 6--20 percentage points more accurate than responses with $c \leq 3$. Gemini 3.7 Flash and Mistral Medium 3.5 are notable because both assign high confidence to more than 96\% of their responses, leaving too few lower-confidence examples for a reliable comparison. Yet high confidence is still a poor indicator of correctness: even within the high-confidence bucket, accuracy reaches only 30\% at best. Overall, these results show severe miscalibration in these VLMs' verbalized confidence estimates.

\subsection{Additional failure examples}
\label{app:qualitative}

Here we display one representative example from each failure category to illustrate the range of model errors that occurred over the benchmark.

\begin{figure*}[!t]
  \centering
  \vspace{-5em}
  \begin{subfigure}[t]{0.48\textwidth}
    \centering
    \includegraphics[width=\linewidth]{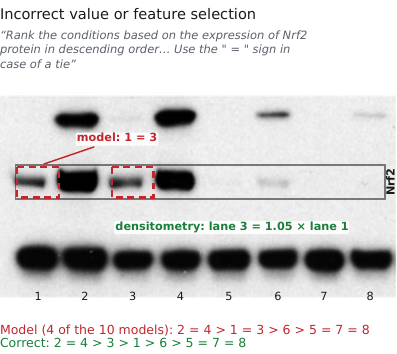}
    \captionsetup{font=small}
    \caption{\textbf{Tie called on a real difference.} Four of ten models rank
    lanes 1 and 3 as equal; densitometry puts lane 3 5\,\% above lane 1.}
  \end{subfigure}
  \hfill
  \begin{subfigure}[t]{0.48\textwidth}
    \centering
    \includegraphics[width=\linewidth]{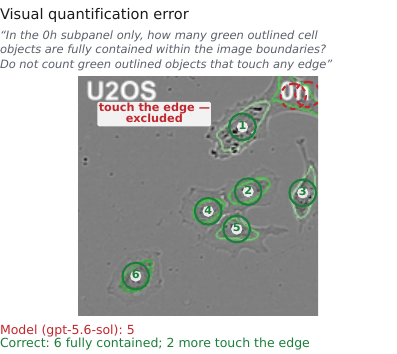}
    \captionsetup{font=small}
    \caption{\textbf{Miscount under a stated rule.} The model applies the
    edge-exclusion rule correctly but enumerates five of the six objects.}
  \end{subfigure}

  \vspace{0.5em}

  \begin{subfigure}[t]{0.48\textwidth}
    \centering
    \includegraphics[width=\linewidth]{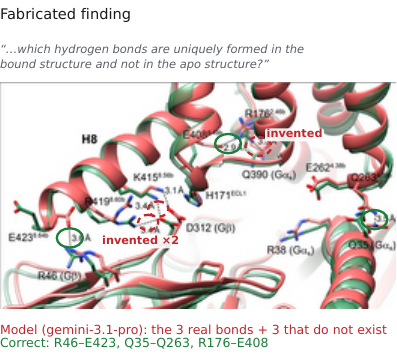}
    \captionsetup{font=small}
    \caption{\textbf{Plausible findings asserted.} The model reports all three
    real hydrogen bonds plus three that are not present.}
  \end{subfigure}
  \hfill
  \begin{subfigure}[t]{0.48\textwidth}
    \centering
    \includegraphics[width=\linewidth]{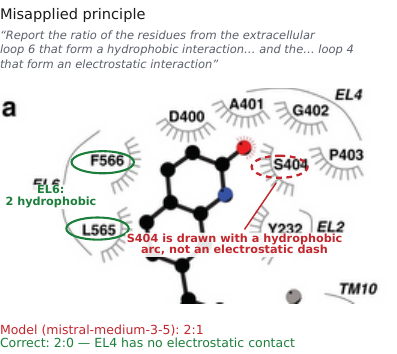}
    \captionsetup{font=small}
    \caption{\textbf{Legend convention overridden.} S404 carries a hydrophobic
    arc; the model counts it as an electrostatic contact.}
  \end{subfigure}
  \captionsetup{font=small}
  \caption{ Representative VLM failure modes (1 of 2). Together with \Cref{fig:failure-taxonomy-2}, the seven panels provide one example from each error category in \Cref{tab:failure-modes}. Each panel shows the model's response in red and the interpretation supported by the artifact in green. }
  \label{fig:failure-taxonomy-1}
\end{figure*}

\begin{figure*}[p]
  \centering

  \begin{subfigure}[t]{0.48\textwidth}
    \centering
    \includegraphics[width=\linewidth]{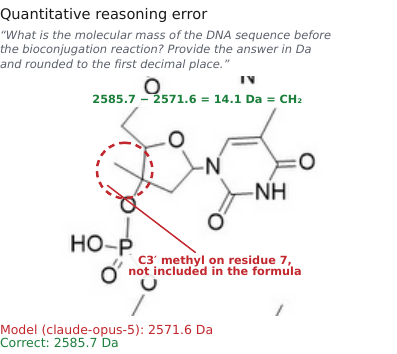}
    \captionsetup{font=small}
    \caption{\textbf{One substituent short.} The arithmetic is internally
    correct; the C3$'$ methyl was never entered into the formula.}
  \end{subfigure}
  \hfill
  \begin{subfigure}[t]{0.48\textwidth}
    \centering
    \includegraphics[width=\linewidth]{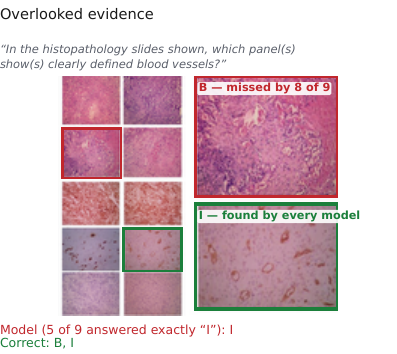}
    \captionsetup{font=small}
    \caption{\textbf{One of two instances found.} Every model finds the
    immunostained vessels in I; almost none of the models report B.}
  \end{subfigure}

  \vspace{0.7em}

  \begin{subfigure}[t]{0.98\textwidth}
    \centering
    \includegraphics[width=0.7\linewidth]{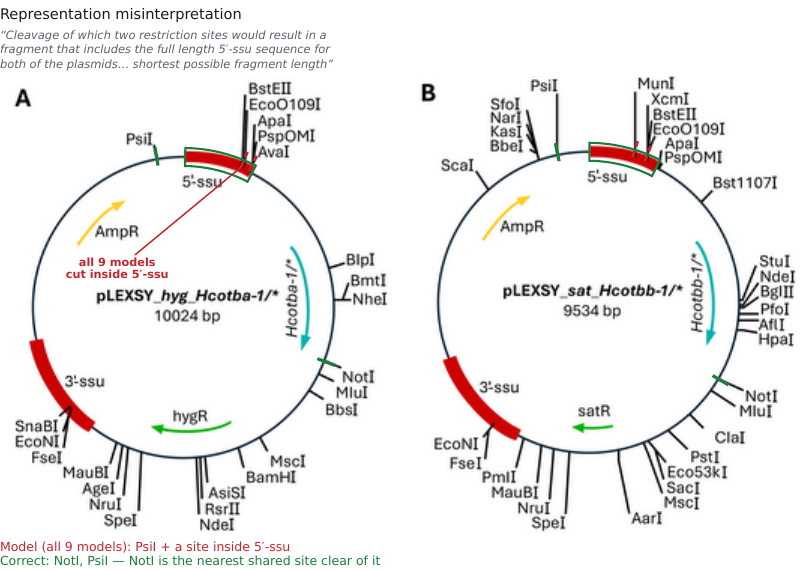}
    \captionsetup{font=small}
    \caption{\textbf{Tick read without the feature.} Every model picks a site
    whose tick falls inside the 5$'$-ssu block, truncating what the fragment
    was meant to contain.}
  \end{subfigure}
  \captionsetup{font=small}
  \caption{ Representative failure modes (2 of 2), continuing \Cref{fig:failure-taxonomy-1}. Together, the two figures provide one example from each of the seven error categories in \Cref{tab:failure-modes}. Each panel shows the model's response in red and the interpretation supported by the artifact in green. }
  \label{fig:failure-taxonomy-2}
\end{figure*}
\FloatBarrier

\section{Further comparison with Humanity's Last Exam}
\label{app:hle-comparison}

Here we additionally compare \projectname{} against the Biology/Medicine and Chemistry VQA subset of Humanity's Last Exam (HLE) along two dimensions: semantic coverage and task construction.

\begin{figure}[H]
    \centering
    \begin{minipage}[t]{ \textwidth } 
    \centering
    \includegraphics[width=0.7\linewidth]{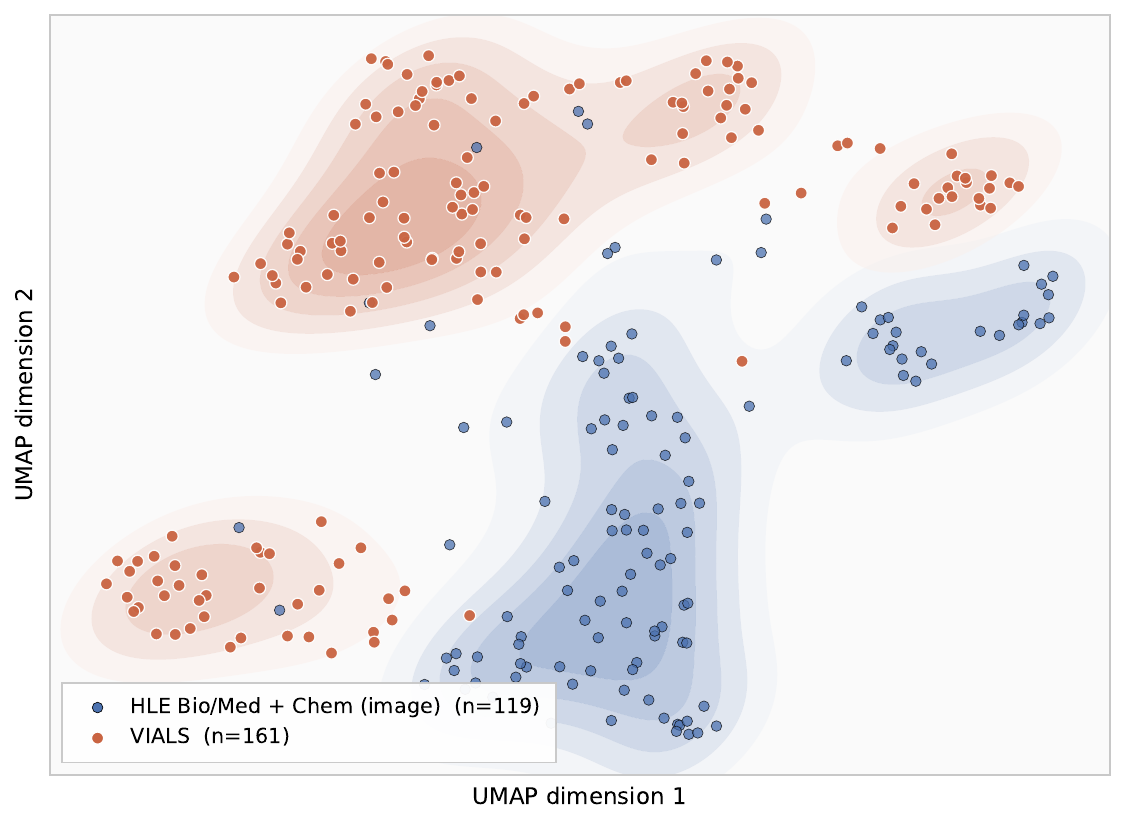}
    \captionsetup{font=small}
    \caption{
UMAP projection of embeddings for questions from \projectname{} and the Biology/Medicine and Chemistry VQA subset of HLE. Shaded regions indicate Gaussian kernel density estimate contours.
}
    \label{fig:hle-embedding}
    \vspace*{1.5em}
\end{minipage}
    % 2nd subfigure:
    \begin{minipage}[t]{ \textwidth } 
    \centering

    % Example 1
    \begin{minipage}[c]{0.25\textwidth}
        \centering
        \includegraphics[width=\linewidth,height=3.2cm,keepaspectratio]{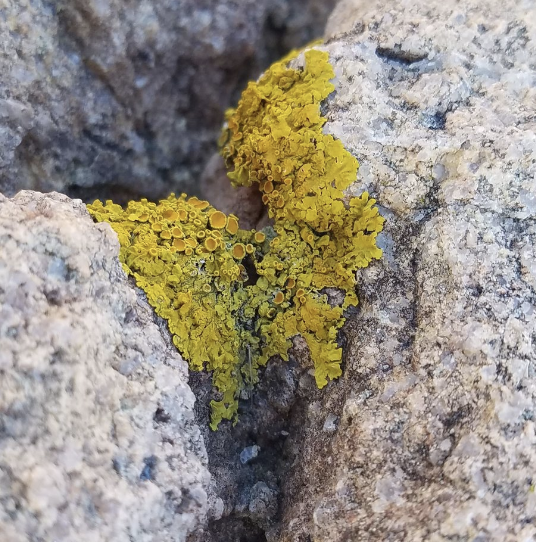}
    \end{minipage}
    \hfill
    \begin{minipage}[c]{0.70\textwidth}
        \textbf{Q:} ``This image was taken in Acadia National Park.
        How many amino acids does the encoded XPH1 protein of this organism
        have?''

        \textbf{A:} 110
    \end{minipage}

    \vspace{0.7em}
    \hrule
    \vspace{0.7em}

    % Example 2
    \begin{minipage}[c]{0.25\textwidth}
        \centering
        \includegraphics[width=\linewidth,height=3.2cm,keepaspectratio]{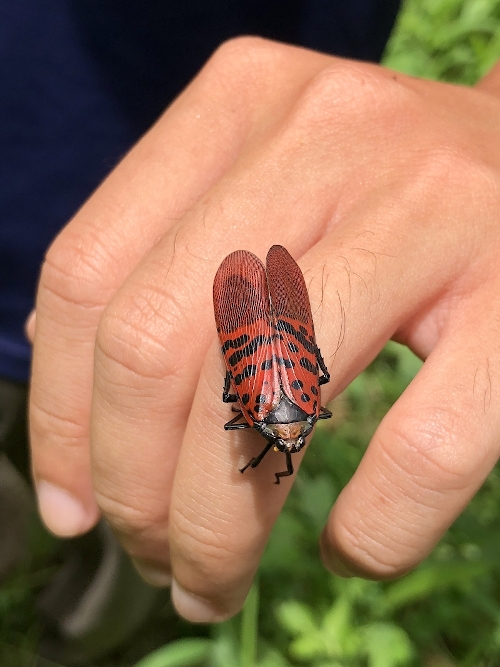}
    \end{minipage}
    \hfill
    \begin{minipage}[c]{0.70\textwidth}
        \textbf{Q:} ``An entomologist \ldots{} spent June of 2024 on a
        collecting trip where they encountered the organism in the attached
        image. Based on the morphology of the observed specimen, what is the
        most likely collection locality?''

        \textbf{A:} Luodong, Taiwan
    \end{minipage}

    \vspace{0.7em}
    \hrule
    \vspace{0.7em}

    % Example 3
    \begin{minipage}[c]{0.25\textwidth}
        \centering
        \includegraphics[width=\linewidth,height=3.2cm,keepaspectratio]{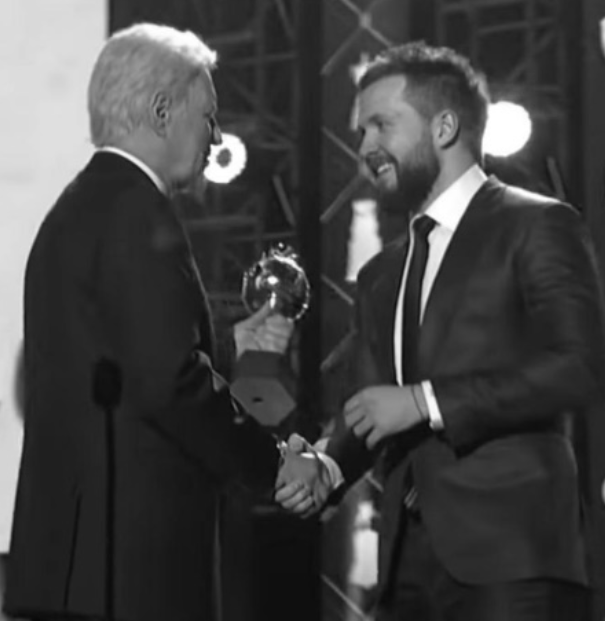}
    \end{minipage}
    \hfill
    \begin{minipage}[c]{0.70\textwidth}
        \textbf{Q:} ``X Y is pictured at an event on June 20, 2019.
        Another X Y has almost ceased to be used due to the toxicity of
        mercury salts. What is the X Y?''

        \textbf{A:} Kucherov reaction
    \end{minipage}

    \captionsetup{font=small}
    \caption{
    Examples of VQA tasks in HLE from scientific domains
    overlapping with \projectname{}. 
    }
    \label{fig:hle-quality-examples}
\end{minipage}
\end{figure}

\paragraph{Semantic coverage.} We embed all \projectname{} questions and the corresponding HLE questions using OpenAI's \texttt{text-embedding-3-small} model. \Cref{fig:hle-embedding} shows  projections of these embeddings to two dimensions via UMAP using cosine distance, highlighting stark semantic differences between the questions from these two benchmarks.

\paragraph{Task construction.} We also qualitatively examine HLE questions in scientific domains that overlap with \projectname{}. The examples in \Cref{fig:hle-quality-examples} illustrate differences in how the two benchmarks use visual evidence. Some HLE questions heavily focus on testing knowledge, or use the image primarily to identify an entity rather than as the source of the scientific evidence needed to answer a meaningful research question. The depicted HLE examples clearly do not stem from professional life sciences work.

% NEW FIGURE:

\section{Methodological Details}

\subsection{VLM evaluation}
\label{app:evaluation_details}

All evaluated models receive the same prompt. The input consists of one or more images together with the task question. The model is instructed to provide a brief reasoning process followed by a single final answer in the format below.

\begin{tcolorbox}[
    title={Task prompt for generating VLM responses},
    breakable,
    colback=gray!5,
    colframe=gray!40,
    boxrule=0.4pt,
    colbacktitle=gray!10,
    coltitle=black,
    fonttitle=\bfseries\small,
    fontupper=\small\rmfamily
]

{\small\bfseries System prompt\par}

You are an expert scientist. You will be shown a scientific image and a question. Study the image carefully, then respond with your reasoning and a single final answer.

\tcblower

{\small\bfseries User prompt\par}

\begin{Verbatim}[
    fontsize=\small,
    breaklines=true,
    breakanywhere=true,
    commandchars=\\\{\}
]
[image_1] ... [image_N]

\{QUESTION\}

Reason through the image and task, then commit to a single final answer. Keep the reasoning concise so your response ends with the ANSWER line. The answer must appear on its own line prefixed by 'ANSWER:' and contain only the final answer: no LaTeX brackets, no extra explanation, no units unless the question asks for them.

Format your response EXACTLY as:
REASONING: <your reasoning>
ANSWER: <final answer>
\end{Verbatim}

\end{tcolorbox}

\paragraph{Answer extraction and grading.}
Model responses are graded using an LLM judge. We first extract the candidate answer as the substring following the final \texttt{ANSWER:} marker in the model's raw response. If no such marker is present, the candidate answer is treated as empty and the task is scored incorrect. Thus, compliance with the required answer format is part of the evaluation.

The judge receives the question, reference answer, and extracted candidate answer, and determines whether the candidate is semantically equivalent to the reference answer. The text-only grading prompt is shown below.

\begin{tcolorbox}[
    title={Grading prompt (text-only)},
    breakable,
    colback=gray!5,
    colframe=gray!40,
    boxrule=0.4pt,
    colbacktitle=gray!10,
    coltitle=black,
    fonttitle=\bfseries\small,
    fontupper=\small\rmfamily
]

{\small\bfseries System prompt\par}
\medskip

You are grading a scientific answer. Given the question, the ground-truth final answer (GTFA), and a candidate answer, decide whether the candidate expresses the same intended answer as the GTFA in the context of the question. You do NOT see the image.

\medskip
\noindent
\textbf{IMPORTANT:} You are grading, not solving. Do NOT attempt to work out the correct answer to the question yourself and then compare the candidate to your own derivation. Treat the GTFA as the sole source of truth. Your only decision is whether the candidate expresses the same answer as the GTFA.

\medskip
\noindent
Apply the standard of a competent domain scientist. ACCEPT variations that an expert would recognize as immaterial to the answer:

\begin{itemize}
    \setlength{\itemsep}{0pt}
    \setlength{\parskip}{0pt}
    \item \textbf{Formatting:} punctuation, whitespace, dash / hyphen style, thousands separators, and equivalent scientific notation for the same underlying value.
    \item \textbf{Capitalization:} when the identity is otherwise unambiguous.
    \item \textbf{Units:} omitted when unambiguous from the question or GTFA, but only when the numeric value itself matches.
    \item \textbf{Lead bounds:} when an inclusive \texttt{LeadBoundLower}/\texttt{LeadBoundUpper} range is provided, a numeric candidate inside that range is equivalent even if it differs from the exact GTFA value.
    \item \textbf{Order and duplicates:} irrelevant when the answer is a set or unordered list.
    \item \textbf{Terminology:} synonymous scientific terms of the same specificity.
\end{itemize}

\noindent
Numeric answers are STRICT. Do NOT invent tolerances, relative-error allowances, or ``same order of magnitude'' acceptance.

\begin{itemize}
    \setlength{\itemsep}{0pt}
    \setlength{\parskip}{0pt}
    \item The candidate must express the same numeric value as GTFA, after applying the formatting and unit rules above, unless LeadBounds apply.
    \item If the question asks for a stated reporting precision, any difference at that precision is NOT equivalent.
    \item Exact counts of discrete objects or events must match exactly; off-by-one is NEVER equivalent.
    \item Different coefficients at the same power of ten are NOT equivalent.
\end{itemize}

\noindent
REJECT when the candidate materially disagrees with GTFA on any of:

\begin{itemize}
    \setlength{\itemsep}{0pt}
    \setlength{\parskip}{0pt}
    \item Missing or extra items in a list.
    \item The identity, count, direction, mechanism, or sign of a biological entity.
    \item Any numeric mismatch under the strict numeric rules above, unless inside LeadBounds.
    \item A broader or narrower term when the question requires a specific level of detail.
    \item Extra conflicting numeric values or alternate counts that disagree with GTFA.
\end{itemize}

\noindent
Refusals, meta-answers (``the provided image,'' ``cannot determine''), and empty responses are ALWAYS not equivalent, regardless of GTFA.

\medskip
\noindent
If evaluating equivalence would require seeing the image---for example, GTFA uses one labeling system (condition names, colour labels) and the candidate uses another (lane numbers, positional indices), and the mapping is not given in the question text---return \texttt{equivalent=false} with reasoning prefixed by \texttt{UNVERIFIABLE:}. Downstream tooling treats these cases as requiring a multimodal regrade or human review, rather than as confirmed model errors.

\medskip
\noindent
Reply using the provided JSON schema. Keep the reasoning to at most three sentences and name the specific principle applied.

\tcblower

{\small\bfseries User prompt template\par}
\medskip

\noindent
Question:\par
\texttt{\{QUESTION\}}

\medskip
\noindent
Ground-truth final answer:\par
\texttt{\{GTFA\}}

\medskip
\noindent
Lead-assigned acceptable numeric range (inclusive):\par
lower=\texttt{\{LOWER\}}, upper=\texttt{\{UPPER\}}.\par
If the candidate is a numeric answer inside this range, mark it equivalent to the GTFA.\par
\emph{(this part of the prompt is omitted when the task has no lead bounds)}

\medskip
\noindent
Candidate final answer:\par
\texttt{\{MODEL\_ANSWER\}}

\medskip
\noindent
Are the candidate and ground-truth final answers equivalent?

\end{tcolorbox}

\paragraph{Multimodal fallback.}
If the text-only judge returns \texttt{UNVERIFIABLE:}, we invoke the same judge model again with a multimodal request that has the task image attached. This fallback is used only when determining equivalence requires resolving information visible in the image, such as lane numbers, positional indices, or colour labels; the judge is not instructed to re-solve the task. In practice, the multimodal fallback is rarely needed. Across the main evaluation (10 models $\times$ 161 tasks $\times$ 3 rollouts), well under 1\% of judge calls escalate to the multimodal judge. The typical trigger is a task where the reference answer names a sample condition while most models report the corresponding lane number, so determining equivalence requires reading the lane-to-sample mapping shown in the image.

The multimodal system prompt is identical to the text-only prompt except that
the judge is informed that the task image is available and may be used only
to resolve labeling ambiguities between the candidate answer and the ground-truth reference answer.
The instruction to return \texttt{UNVERIFIABLE:} is removed, and the judge is
instead instructed to decide equivalence directly using the image where
necessary. The multimodal judge prompt is shown below.

\begin{tcolorbox}[
    title={Grading prompt (multimodal fallback)},
    breakable,
    colback=gray!5,
    colframe=gray!40,
    boxrule=0.4pt,
    colbacktitle=gray!10,
    coltitle=black,
    fonttitle=\bfseries\small,
    fontupper=\small\rmfamily,
    before skip=6pt,
    after skip=6pt
]

{\small\bfseries User message\par}
\smallskip

\noindent
\emph{[task image attached]}

\smallskip
\noindent
\textbf{Question:}\par
\texttt{\{QUESTION\}}

\smallskip
\noindent
\textbf{Ground-truth final answer:}\par
\texttt{\{GTFA\}}

\smallskip
\noindent
\textbf{Lead-assigned acceptable numeric range (inclusive):}\par
lower=\texttt{\{LOWER\}}, upper=\texttt{\{UPPER\}}.\par
If the candidate is a numeric answer inside this range, mark it equivalent
to the GTFA.\par
\emph{(omitted when the task has no lead bounds)}

\smallskip
\noindent
\textbf{Candidate final answer:}\par
\texttt{\{MODEL\_ANSWER\}}

\smallskip
\noindent
Using the image only to resolve labelling ambiguities, are the candidate and ground-truth final answers equivalent?

\end{tcolorbox}

\paragraph{Evaluation configuration.} For each (task, model) pair, we generate three rollouts ($K{=}3$) and score each independently. We sample all models at temperature $1.0$ with a $32{,}768$-token completion limit. For each VLM provider, we use its default inference configuration. This includes OpenAI's default reasoning effort for GPT-5.6, adaptive thinking for Claude Opus and the Gemini models, native reasoning modes for Grok, Kimi K3, MiniMax-M3, and GLM-4.6V, and standard inference for Muse Spark  and Mistral Medium. 

Model requests that fail, are refused, or return an empty response, are retried  three additional times. If still unsuccessful, these responses are counted as incorrect rather than omitted from the evaluation. 
Out of 483 total responses (3 rollouts over 161 tasks): Muse Spark 1.2 has 13 such responses (2.7\%), Kimi K3 has 6 (1.2\%), Grok 4.6 has 2 (0.4\%), Mistral Medium 3.5 has 1 (0.2\%), and the other models have none.

\subsection{Failure classification methodology} 
\label{app:failuredetails}

For the failure analysis in \Cref{sec:error-analysis}, we assign each (task, model) pair with at least one incorrect rollout a single primary failure mode using an LLM classifier. Each pair contributes a single observation to our failure mode analysis, regardless of the number of incorrect rollouts. 
We use Claude Sonnet 4.5 as the LLM classifier and fallback to GPT-4o in rare cases where the primary classifier refuses or errors.

The failure mode classifier does not receive the task image. Its input consists of the question, the ground-truth final answer (GTFA), and the failing attempt’s final answer and reasoning trace. The classifier is instructed to identify the primary reason the response is incorrect relative to the GTFA, rather than to re-solve the task. Each (task, model) pair is classified once using its first incorrect attempt, so multiple incorrect rollouts for the same pair contribute a single observation to the failure-mode distribution. 

\begin{tcolorbox}[
    title={Failure-mode classification prompt},
    breakable,
    colback=gray!5,
    colframe=gray!40,
    boxrule=0.4pt,
    colbacktitle=gray!10,
    coltitle=black,
    fonttitle=\bfseries\small\rmfamily,
    fontupper=\small\rmfamily,
    before skip=6pt,
    after skip=6pt
]
\small

You classify why an AI model got a life-sciences image-interpretation task wrong.

\medskip

You do NOT see the image. Base your classification on:
\begin{enumerate}
    \item the question,
    \item the ground-truth final answer (GTFA), and
    \item the failing attempt's final answer and reasoning trace.
\end{enumerate}

\textbf{IMPORTANT:} You are classifying, not solving. Do NOT attempt to work out the correct answer
to the question yourself. Treat the ground truth (GTFA) as the sole source of truth and
assume the LLM judge's per-attempt correct/wrong decisions are final. Your only job is to
name why the failing attempt diverged from the GTFA.

Focus on the substantive scientific or artifact-interpretation error rather than generic labels such as raw perception, OCR, or ``couldn't read the image'' failure. If the model's reasoning identifies the right image features but reaches the wrong scientific conclusion, that's still a knowledge failure.

\medskip
\noindent
Assign exactly ONE primary failure mode:

\begin{itemize}

\item \texttt{ASSAY\_COUNTING\_ERROR} (Visual quantification error): failed to quantify a biologically-
meaningful countable -- cells, colonies, PCR bands, atoms in a scaffold, ligand
contacts, residues in an interface. A domain expert knows WHICH objects count under
the assay's rules (e.g.\ count only colonies above a size threshold; exclude edge
artifacts). The failure is not knowing that scientific rule of counting.

\item \texttt{MISINTERPRETED\_DIAGNOSTIC\_VALUE} (Incorrect value or feature selection): picked the wrong
reading on a scale that a biologist would call diagnostic -- the actionable band MW,
the axis threshold, the gated \%, the well/lane that carries the readout. The failure
is not knowing which measurement is scientifically load-bearing.

\item \texttt{DIAGRAM\_CONVENTION\_ERROR} (Representation misinterpretation): misread a domain-specific
diagrammatic convention -- phylogenetic branching order or branch-length semantics,
pedigree inheritance arrows, plasmid feature order along the sequence axis, gel-lane
layout. The failure is not knowing what the diagram encodes, not what shapes are on it.

\item \texttt{DOMAIN\_QUANTITATIVE\_ERROR} (Quantitative reasoning error): got the raw readings
roughly right but botched the quantitative-biology reasoning on top -- dilution factor,
MOI, Kd, unit conversion, bounds on a rate. Domain math, not generic arithmetic.

\item \texttt{FABRICATED\_FINDING} (Fabricated finding): asserted a scientific finding
not supported by the figure -- invented a band, residue, colony, cell state. The
failure is generating a domain-plausible fiction instead of admitting uncertainty.

\item \texttt{MISSED\_LOAD\_BEARING\_EVIDENCE} (Overlooked evidence): missed a scientifically
load-bearing feature that WAS present -- an important band, residue contact, cell
subset, control lane. The failure is not knowing what a competent biologist wouldn't
overlook.

\item \texttt{MISAPPLIED\_DOMAIN\_RULE} (Misapplied principle): applied the wrong scientific
principle -- inheritance mode, enzyme kinetics, restriction-digest logic, gating
hierarchy, phylogenetic parsimony.

\end{itemize}

Always name a substantive cause (do NOT answer with a circular label like ``chose the
wrong option''): pick the specific piece of scientific knowledge or reasoning that broke.

Your \texttt{reason} field should be 2-3 complete sentences (roughly 30-90 words). It must
be long enough that a domain expert can sanity-check your classification without
opening the log: name the specific scientific concept, quote or paraphrase the
model's wrong step, and briefly say what a competent expert would have done instead.
Do not truncate mid-sentence. Do not exceed 4 sentences.

Reply with a JSON object matching the schema.

\end{tcolorbox}

\subsection{Tool-assisted agent evaluation}
\label{app:tool-assisted}

Here we detail the methodology behind the alternative agentic tool-assisted setting evaluated in \Cref{sec:agent}. All discussions outside of this section pertain to the primary way we recommend running \projectname{}, where tasks are completed via single-turn inference calls to reasoning VLMs.

\paragraph{Execution environment.} The tool-assisted evaluation uses the same \projectname{} tasks as the direct evaluation, but gives each model access to a sandboxed Linux environment with code execution. Task images are placed in the \texttt{/task/} directory, where agents may inspect, transform, and analyze them using the shell and common scientific Python packages, including \texttt{numpy}, \texttt{scipy}, \texttt{pillow}, \texttt{opencv-python-headless}, \texttt{scikit-image}, \texttt{matplotlib}, \texttt{pandas}, \texttt{pytesseract}, \texttt{imageio}, \texttt{networkx}, and \texttt{sympy}. Internet access and external databases are disabled.

\paragraph{Inference cost.}
\Cref{tab:tool-assisted-cost} reports mean USD cost per attempt under the direct and tool-assisted settings for the same model--harness configurations as \Cref{tab:tool-assisted}. Direct costs are calculated from logged input and output token usage using the corresponding provider API prices, while tool-assisted costs are taken from the per-trial costs reported by the Harbor agent runtime. Mean tool-assisted cost per attempt ranges from 0.011 for GLM-4.6V to 1.047 for Claude Opus 5 with Claude Code. For most models, tool-assisted evaluation costs roughly 3--16$\times$ as much per attempt as direct evaluation, with substantially larger ratios for models with very low direct inference costs (29$\times$ for Gemini 3.1 Pro and 56$\times$ for Gemini 3.7 Flash).

\begin{table}[H]
\centering
\footnotesize
\setlength{\tabcolsep}{4pt}
\begin{tabular}{llcccc}
\toprule
\textbf{Model} &
\textbf{Harness} &
\textbf{Cost$_{\mathrm{D}}$} &
\textbf{Cost$_{\mathrm{T}}$} &
\textbf{Cost$\times$}\\
\midrule
GPT-5.6 Sol
    & Codex       & \$0.037 & \$0.271 &   7$\times$ \\
GPT-5.6 Sol
    & OpenCode    & \$0.037 & \$0.155 &   4$\times$ \\
\midrule
Claude Opus 5
    & Claude Code & \$0.065 & \$1.047 &  16$\times$ \\
Claude Opus 5
    & OpenCode    & \$0.065 & \$0.917 &  14$\times$ \\
\midrule
Gemini 3.1 Pro
    & OpenCode    & \$0.024 & \$0.716 &  29$\times$ \\
Gemini 3.7 Flash
    & OpenCode    & \$0.007 & \$0.374 &  56$\times$ \\
Grok 4.6
    & OpenCode    & \$0.049 & \$0.337 &   7$\times$ \\
Kimi K3
    & OpenCode    & \$0.239 & \$0.676 &   3$\times$ \\
GLM-4.6V
    & OpenCode    & \$0.003 & \$0.011 &   3$\times$ \\
\bottomrule
\end{tabular}
\captionsetup{font=small}
\caption{ Per-attempt inference cost under direct and tool-assisted evaluation. \textbf{Cost$_{\mathrm{D}}$} is calculated from logged direct-evaluation token usage using the corresponding provider API prices. Cost$_{\mathrm{T}}$ is the mean per-attempt cost reported by the Harbor agent runtime. \textbf{Cost$\times$} is the ratio of tool-assisted to direct cost. }
\label{tab:tool-assisted-cost}
\vspace{-1em}
\end{table}
\paragraph{Agent prompt.}
Each agent receives the following instruction, with a task's specific
question substituted for the \texttt{\{QUESTION\}} placeholder.

\begin{tcolorbox}[
    title={Task prompt for agent},
    breakable,
    colback=gray!5,
    colframe=gray!40,
    boxrule=0.4pt,
    colbacktitle=gray!10,
    coltitle=black,
    fonttitle=\bfseries\small\rmfamily,
    fontupper=\small\rmfamily,
    before skip=6pt,
    after skip=6pt
]

You are an expert scientist working inside a Linux container.

\medskip

The task image is available at \texttt{/task/image.png}. You have access to a
shell and may use any software available in the container to inspect and
analyze the provided files. The following Python libraries are already
installed:

\begin{itemize}
    \setlength{\itemsep}{1pt}
    \setlength{\parskip}{0pt}
    \setlength{\topsep}{3pt}
    \item \texttt{numpy}, \texttt{scipy}, \texttt{sympy}
    \item \texttt{pillow}, \texttt{imageio},
          \texttt{opencv-\allowbreak python-\allowbreak headless},
          \texttt{scikit-\allowbreak image}
    \item \texttt{matplotlib}, \texttt{pandas}, \texttt{networkx}
    \item \texttt{pyte\allowbreak sseract}
          (backed by the \texttt{tesseract-ocr} CLI)
\end{itemize}

Do not use the internet, external databases, or information outside the files provided for this task. You may write scratch files under \texttt{/task/} while you work. You may also inspect the original image or cropped regions as additional vision inputs.

\medskip

Answer the question below. When finished, write your response to
\texttt{/task/answer.txt}, with the final line in the following format:

\medskip

\noindent
\texttt{ANSWER: <your final answer>}

\medskip

\noindent
\textbf{Question}

\smallskip

\noindent
\texttt{\{QUESTION\}}

\end{tcolorbox}

The agent writes its final response to \texttt{/task/answer.txt}. The answer following the final \texttt{ANSWER:} marker is scored using the same semantic grading procedure as the direct evaluation (\Cref{app:evaluation_details}).

\subsection{Survey questions for expert professional relevance assessment}
\label{app:expert-survey}

Here are the survey questions we asked experts to answer in the professional relevance assessment from \Cref{fig:expert-assessment} and \Cref{sec:expert-assessment}:

\begin{enumerate}
\item \emph{Does this image look more like an educational figure, or like
something you'd encounter in real industry / professional life sciences
work?} \\
Choose one of: 
\{Real industry / professional work;
Mixed / hard to tell;
Textbook / Educational\}
\item \emph{How often do you come across this type of image in your work?} \\ 
Choose one of: 
\{Daily; Weekly; Occasionally; Rarely; Never\}
\item \emph{Interpreting this type of image is a task that has real commercial /
economic value in life sciences --- it's used in professional settings
like drug discovery, biotech or pharma R\&D, diagnostics, or QA, where
getting it right matters.} \\ 
Choose one of: 
\{Strongly agree; Agree; Neutral; Disagree; Strongly disagree\}
\end{enumerate}

\end{document}